\documentclass[lettersize,journal]{IEEEtran}
\usepackage{amsmath,amsfonts}
\usepackage{algorithmic}
\usepackage{algorithm}
\usepackage{array}
\usepackage{textcomp}
\usepackage{stfloats}
\usepackage{url}
\usepackage{verbatim}
\usepackage{graphicx}
\usepackage{cite}

\usepackage{amssymb}
\usepackage{xcolor}
\usepackage{bm}
\usepackage{latexsym}
\usepackage[hang,small,bf]{caption}
\usepackage[subrefformat=parens]{subcaption}
\begin{document}

\title{Multi-Output Gaussian Processes for Graph-Structured Data}

\author{Ayano Nakai-Kasai,~\IEEEmembership{Member,~IEEE,} 
        \thanks{A. Nakai-Kasai and T. Wadayama are with Graduate School of Engineering, Nagoya Institute of Technology, Gokiso, Nagoya, Aichi 466-8555, Japan,}
      and Tadashi Wadayama,~\IEEEmembership{Member,~IEEE} 
}

\markboth{Journal of \LaTeX\ Class Files,~Vol.~14, No.~8, August~2021}%
{Shell \MakeLowercase{\textit{et al.}}: A Sample Article Using IEEEtran.cls for IEEE Journals}


\maketitle

\begin{abstract}
    Graph-structured data is a type of data to be obtained associated with a graph structure where vertices and edges describe some kind of data correlation.
    This paper proposes a regression method on graph-structured data, 
    which is based on multi-output Gaussian processes (MOGP), to capture both the correlation between vertices and the correlation between associated data.
    The proposed formulation is built on the definition of MOGP. 
    This allows it to be applied to a wide range of data configurations and scenarios. 
    Moreover, it has high expressive capability due to its flexibility in kernel design.
    It includes existing methods of Gaussian processes for graph-structured data as special cases 
    and is possible to remove restrictions on data configurations, model selection, and inference scenarios in the existing methods.
    The performance of extensions achievable by the proposed formulation is evaluated through computer experiments with synthetic and real data.
\end{abstract}


\section{Introduction}
\subsection{Background}



\IEEEPARstart{G}{aussian} process regression is a nonparametric supervised learning method that can handle nonlinear predictions with only a few or several parameters \cite{GPML}.
The flexibility of data configurations, inference scenarios, and the design of the covariance function, which appears as a kernel, is high. 
It has many applications due to its compatibility with geographical and location-related data \cite{Cressie,Stein,Deisenroth,Liutut}.
Because Gaussian process regression is a Bayesian method, it is possible to obtain the prediction uncertainty as a covariance matrix of the Gaussian distribution, 
which can be used for Bayesian optimization \cite{Contal}.
Theoretical support for its generalization capability \cite{Kanagawa} and the fact that it is a mathematical alternative to neural networks \cite{Neal} 
also encourages the use of the method.

The application of Gaussian process regression to graph-structured data and graph signals, 
which describes the correlation between data points as vertices and edges, 
is also being considered \cite{Dong,Smola}.
Graph-structured data can be, for example, brain function networks \cite{fmri}, traffic networks \cite{traffic}, or data observed on a sensor network \cite{sensornetwork}.
Generalization performance can be enhanced by incorporating the graph structure into the Gaussian process kernel. 
This is achieved by using the vertex information (representing graph signals) and edge information (representing connections between vertices) to define the kernel correlations.
Standard Gaussian process regression is called single-output Gaussian process (SOGP) regression.
There is a method called multi-output Gaussian process (MOGP) regression \cite{MOGP,GPML,Cressie} that is an extension of SOGP.
Along with correlations in the input data, 
a kind of higher-order correlation in the output can be incorporated,
which is useful for data where there is a obvious correlation among the outputs.
Graph-structured data meet this condition and have great potential for use 
because the graph structure naturally gives useful correlations among the vertices.

Some regression methods \cite{Venkitaraman,Zhi} proposed for graph-structured data are equivalent to MOGP regression 
and have been successfully applied to synthetic and real data such as electroencephalography signals.
These methods are formulated under an observation model 
in which the inputs are identical for all vertices 
and the signals for all vertices are obtained together.
In other words, the model cannot handle signals that are observed in different regions for each vertex, 
and vertices with partially missing data \cite{missing} should be eliminated from the model, 
which does not cover realistic data collection scenarios.
Furthermore, kernel is a key model determinant in Gaussian process regression, but 
these methods also advocate the choice of a particular form of kernel formulated as a graph filter.
This limitation of the model can lead to a lack of flexibility 
because, in Gaussian processes, kernel selection and kernel combinations based on prior information of data 
have a significant impact on generalization performance.
For example, when data have both trend and periodicity, such as temperature data \cite{GPML}, 
or when there are multiple data series behind the data that affect different graphical scales, 
such as seismic waves \cite{seismic} or traffic congestion \cite{Boro}, 
modeling by combining multiple types of kernels is appropriate.
If MOGP can be formulated to allow removing such restrictions and handling of a variety of problems, 
it would be a powerful method that is flexible, simple (not requiring too many parameters), and with interpretability.
In other words, a more general formulation covering a wider range of applications is required 
to take full advantage of the benefits of MOGP.

In this paper, we return to the definition of MOGP 
and provide a generalized formulation for regression problems of data observed on a graph.
The application of MOGP regression to graph-structured data is expected to improve performance by incorporating graph structure into a simple model with high expressive capability.
All of the benefits of using MOGP can only be realized by formulating the method from the definition.
This allows us to cover a wider range of data collection scenarios and enhance the flexibility of the kernel design.
In particular, we will present important extensions of the method, 
including the extensions of input data configurations and scenarios not envisioned by existing methods, 
as well as the full exploitation of graph structures related to input data.
Detailed kernel design methods that incorporate graph structures will also be discussed.
The range of possible applications is so broad that it is difficult to present them all in this paper, 
but there is no doubt that the proposed formulation can be applied to various applications.

\subsection{Related Work}
Parametric methods have been used for prediction problems on graph-structured data regardless of GP \cite{Bronstein,Phillips}. 
GNN has also an applicability to prediction on graph-structured data \cite{Henaff}.
The number of parameters is usually several hundred or more 
and training on a large amount of training data is required.

A number of kernels that incorporate graph structure in SOGP regression have been proposed \cite{Smola,Boro,Zhi,Wilson,MOGP,Venkitaraman} 
and are summarized in \cite{Smola,Zhi2}.
It has been shown that they can be derived in terms of regularization in kernel regression in the graph spectral domain.
They function as graph filters, for example, low-pass filters.
They perform well when the data structure is strongly connected to the graph structure, 
but they may not be able to absorb the characteristics of the background data themselves, 
and are not designed to follow data characteristics such as obvious trends or periodicity.

Venkitaraman et al. and Zhi et al. proposed method to incorporate both graph structure and data properties 
from the perspective of graph filters \cite{Venkitaraman,Zhi}. 
The graph kernels for SOGP introduced above \cite{Smola} or polynomial graph filters \cite{Zhi,Zhi2} can be incorporated into the methods as the graph filters.
It is shown in the papers that these methods can be represented as a kind of MOGPs.

MOGP pioneered in the field of geostatistics and known as cokriging \cite{ICM}.
The typical regression models are the linear model of coregionalization and its simplified version, intrinsic coregionalization model (ICM).
The ICM coincides with MOGP regression with parameter vectors, which is introduced in Sect.~\ref{sec:experiments}.
The model is regarded as the MOGP regression that does not explicitly introduce a graph structure.


\subsection{Contributions}
\begin{table*}[tb]
    \centering
    \caption{Comparison of the proposed formulation with conventional approaches. 
    Limited in SOGP means that the graph information is available by using graph kernels but the characteristics of the data cannot be taken into account simultaneously.}
  \label{tab:contributions}
    \begin{tabular}{|cc|ccccc|}
      \hline
      \multicolumn{2}{|c|}{Gaussian process} & Graph Structure & Data Configuration & Data Scenario & Kernel Selection & Inference Scenario \\
      \hline\hline
      \multicolumn{2}{|c|}{SOGP} & Available, Limited & Iso/Heterotopic & Symmetric/Asymmetric & Limited & Flexible \\
      \hline
      & ICM & Unavailable & Iso/Heterotopic & Symmetric/Asymmetric & Specific & Flexible \\
      MOGP & for Graph \cite{Venkitaraman,Zhi} & Available & Isotopic & Symmetric & Specific & Limited \\
      & \textbf{for Graph, This paper} & Available & Iso/Heterotopic & Symmetric/Asymmetric & Flexible & Flexible \\
      \hline\hline
      \multicolumn{2}{|c|}{Corresponding Section} & \ref{sec:formulation} & \ref{sec:input} & \ref{sec:input} & \ref{sec:kernel}, \ref{sec:conv} & \ref{sec:semi} \\
      \hline
    \end{tabular}
\end{table*}
The contributions of this paper are the following. 
Table~\ref{tab:contributions} summarizes the differences between conventional approaches and the proposed formulation, and shows the corresponding section numbers in the paper.
\begin{itemize}
    \item This paper derives the MOGP regression method for graph-structured data from the definition of MOGP (Sect.~\ref{sec:formulation}).
      This formulation can introduce not only data characteristics but also graph information 
      that cannot be fully captured by SOGP or ICM.
    \item This paper introduces some specific extensions that cannot be covered by the existing MOGP regression methods for graph-structured data \cite{Venkitaraman,Zhi}. 
      The proposed formulation removes constraints on data configurations, kernel selection, and inference scenarios, 
      allowing us to receive the full benefit of a wide range of applicability.
      \begin{itemize}
        \item First, the proposed formulation can extend the configurations and scenarios of input data (Sect. \ref{sec:input}). 
        \item Second, the proposed formulation allows for the adoption of sophisticated kernels (Sects. \ref{sec:kernel} and \ref{sec:conv}). This leads to an improvement in model flexibility.
        \item Third, the proposed formulation can be applied to a wider range of inference scenarios (Sect. \ref{sec:semi}). Graph information can be used to estimate problems such as missing data estimation, where the outputs to be estimated are not available at all vertices.
      \end{itemize}
    \item We propose a novel kernel, graph process convolution, for graph-structured data as one of the extensions (Sect.~\ref{sec:conv}). 
    \item Computer experiments with synthetic and real data corresponding to the above examples are performed to show that the proposed formulation can improve the expressive capability of the model (Sect.~\ref{sec:experiments}).
\end{itemize}

The model has a wide range of potential applications, not limited to the extensions shown in this paper.
On the basis of the proposed formulation, 
we can also expect to see prospects for unsupervised learning and reduction of computational complexity.



\section{Preliminaries}
\label{sec:pre}
\subsection{Notation}
In the rest of the paper, we use the following notation.
Superscript $(\cdot)^{\mathrm{T}}$ denotes the transpose.
The zero vector of size $N$ and identity matrix of size $N\times N$ are represented by $\bm{0}_N$ and $\bm{I}_N$, respectively.
The Euclidean ($\ell_2$) norm is $\|\cdot\|$.
The Gaussian distribution $\mathcal{N}(\bm{0}_N,\bm{\Sigma})$ 
has a mean vector $\bm{0}_N$ and a covariance matrix $\bm{\Sigma}$.
The expectation, trace, and vectorization operators are $\mathbb{E}[\cdot]$, $\mathrm{Tr}[\cdot]$, and $\mathrm{vec}[\cdot]$, respectively.
The diagonal matrix is given by $\mathrm{diag}[\ldots]$ with the diagonal elements shown in square brackets.
The determinant of a matrix $\bm{A}$ is $\mathrm{det}[\bm{A}]$.

\subsection{Gaussian process}
Gaussian process \cite{GPML} is a stochastic process where any finite set of random variables follows a multivariate normal distribution.
For any input $\bm{x} \in\mathcal{Q}\subset\mathbb{R}^D$, the Gaussian process $f: \mathbb{R}^D\to\mathbb{R}$ is described as 
\begin{equation}
  f(\bm{x})\sim\mathcal{GP}(m(\bm{x}),k(\bm{x},\bm{x}')),
  \label{eq:gp}
\end{equation}
where $m(\bm{x})=\mathbb{E}[f(\bm{x})]$ is a mean function 
and $k(\bm{x},\bm{x}')=\mathbb{E}[(f(\bm{x})-m(\bm{x})(f(\bm{x}')-m(\bm{x}')))]$ is a covariance function.
The mean function of Gaussian processes can be assumed to be zero without loss of generality.
The covariance function characterizes the process. 
The choice of covariance function corresponds to model selection in Gaussian processes.
Typical selections are 
squared exponential (SE) function: 
\begin{equation}
	k(\bm{x}, \bm{x}') = v^2 \exp\left(-\frac{r^2}{2\ell}\right),
	\label{eq:se}
\end{equation}
and the Mat\'{e}rn-$\nu$ function: 
\begin{equation}
	k(\bm{x},\bm{x}') = v^2 \frac{2^{1-\nu}}{\Gamma(\nu)}\left(\sqrt{2\nu}\frac{r}{\ell}\right)^\nu K_\nu\left(\sqrt{2\nu}\frac{r}{\ell}\right),
	\label{eq:matern}
\end{equation}
where $r=\|\bm{x}-\bm{x}'\|$ is the Euclidean distance between $\bm{x}\in\mathcal{Q}$ 
and $\bm{x}'\in\mathcal{Q}$, 
where $\nu >0$ is a model parameter for Mat\'{e}rn function, 
where $\Gamma(\cdot)$ and $K_\nu(\cdot)$ are the Gamma function and the modified Bessel function of the second kind \cite{bessel}, respectively, 
and where $v^2>0$ and $\ell>0$ are hyperparameters.

\subsection{Single-Output Gaussian Process Regression}
Consider SOGP regression given a training dataset with $N$ samples, 
$\mathcal{D}=\{(\bm{x}_n,y_n)\in\mathcal{Q}\times\mathbb{R}\}_{n=1,\ldots,N}$.
The data are summarized as $\bm{X} = [\bm{x}_{1},\cdots,\bm{x}_{N}]^{\mathrm{T}} \in \mathbb{R}^{N \times D}$ and $\bm{y} = [y_{1},\cdots,y_{N}]^{\mathrm{T}} \in \mathbb{R}^N$.
The observation model is assumed as follows:
\begin{equation}
  y_n=f(\bm{x}_n)+\epsilon_n.
	\label{eq:obs}
\end{equation}
The regression function $f$ follows the Gaussian process \eqref{eq:gp}.
The residual error $\epsilon_n\in\mathbb{R}$ is assumed to be a white Gaussian noise that follows $\mathcal{N}(0,\sigma^2)$.

The objective is prediction of the output values, $\bm{y}^\ast = [y_1^\ast, \ldots, y_{T}^\ast]^{\mathrm{T}} \in \mathbb{R}^{T}$, 
at $T$ test points $\bm{X}^\ast = [\bm{x}_{1}^\ast,\cdots,\bm{x}_{T}^\ast]^{\mathrm{T}} \in \mathbb{R}^{T \times D}$.
On the assumption that the prior of $f$ is GP,
the joint distribution of $\bm{y}$ and $\bm{y}^\ast$ is given by 
\begin{align}
	& p\left(\left[\begin{array}{cc}\bm{y}\\\bm{y}^\ast\end{array}\right]\right) = \nonumber \\
  & \mathcal{N}\left(\bm{0}_{N+T},\left[\begin{array}{cc}\bm{K}(\bm{X})&\bm{K}(\bm{X},\bm{X}^\ast)\\\bm{K}(\bm{X}^\ast,\bm{X})&\bm{K}(\bm{X}^\ast)\end{array}\right] + \sigma^2\bm{I}_{N+T}\right),
	\label{eq:full}
\end{align}
where the covariance matrix $\bm{K}(\bm{X},\bm{X}')$ is such that $[\bm{K}(\bm{X},\bm{X}')]_{nn'}=k(\bm{x}_n,\bm{x}'_{n'})$ 
and $\bm{K}(\bm{A},\bm{A})$ is shortened to $\bm{K}(\bm{A})$ for any matrix $\bm{A}$.
The predictive distribution of $\bm{y}^\ast$ given $\mathcal{D}$ can be obtained as the conditional Gaussian distribution 
$p(\bm{y}^{\ast} | \bm{X}^{\ast}, \mathcal{D}) = \mathcal{N}\left(\bm{\mu}_{\mathrm{SO}}(\bm{X}^\ast), \bm{\Sigma}_{\mathrm{SO}}(\bm{X}^{\ast})\right)$, where 
\begin{align}
	\bm{\mu}_{\mathrm{SO}}(\bm{X}^\ast) &= \bm{K}(\bm{X}^{\ast},\bm{X})(\bm{K}(\bm{X})+\sigma^2\bm{I}_N)^{-1} \bm{y}, 
	\label{eq:mufull}\\
	\bm{\Sigma}_{\mathrm{SO}}(\bm{X}^{\ast}) &=  \sigma^2\bm{I}_T + \bm{K}(\bm{X}^\ast) \nonumber\\
  &- \bm{K}(\bm{X}^{\ast},\bm{X}) (\bm{K}(\bm{X})+\sigma^2\bm{I}_N)^{-1} \bm{K}(\bm{X},\bm{X}^{\ast}).
	\label{eq:sigma2full}
\end{align}
In other words, the predicted values and the uncertainty 
are obtained as the mean $\bm{\mu}_{\mathrm{SO}}$ and covariance matrix $\bm{\Sigma}_{\mathrm{SO}}$ of the distribution, respectively. 

Hyperparameters including noise variance, such as $\{v^2, \ell, \sigma^2\}$ in the case of \eqref{eq:se},  
are determined via maximizing the following log-marginal likelihood using some optimization method:
\begin{align}
	\log{p(\bm{y}|\bm{X},\bm{\Theta})} &\propto -\frac{1}{2}\Big(\bm{y}^{\mathrm{T}}(\bm{K}(\bm{X})+\sigma^2\bm{I}_N)^{-1}\bm{y} \nonumber \\
  &\quad + \log{\mathrm{det} \left[\bm{K}(\bm{X})+\sigma^2\bm{I}_N\right]}\Big), 
	\label{eq:ml}
\end{align}
where $\bm{\Theta}$ is the set of hyperparameters.

\section{Multi-Output Gaussian Process Regression for Graph-Structured Data}
\label{sec:mogpongraph}

\subsection{Formulation}
\label{sec:formulation}
In this subsection, we introduce the formulation of regression for graph-structured data on the basis of the definition of MOGP.
MOGP considers vector-valued output of Gaussian processes and correlations among the elements of the output \cite{MOGP}.
The proposed regression model is formulated by regarding signals observed on vertices of a graph 
as the vector-valued output.
The correlations among the elements of the output in this case should include both connectivity of the vertices and underlying data distribution.
On the basis of the definition of MOGP, 
it can model relationships between arbitrary inputs and outputs, 
thus covering input data configurations and inference scenarios that cannot by existing methods.
Moreover, the correlation among the elements is represented as kernel so that it allows for flexible models that are not limited to a specific graph filter.

Consider a graph $\mathcal{G}=(\mathcal{V},\mathcal{E})$ with $|\mathcal{V}|=M$ nodes.
Input and output data tied to each node are represented as $\bm{X}_m\in\mathbb{R}^{N_m\times D}$ and $\bm{Y}_m\in\mathbb{R}^{N_m}$.
The data are summarized as $\mathcal{D}_\mathcal{M}=\{\mathcal{D}_1,\ldots,\mathcal{D}_M\}$, 
where $\mathcal{D}_m = \{\bm{X}_m,\bm{Y}_m\} =\{(\bm{x}_{m,n},y_{m,n})\in\mathcal{Q}\times\mathbb{R}\}_{n=1,\ldots,N_m}$ for $m=1,\ldots,M$ 
and we re-difine $N=\sum_{m=1}^M N_m$.

The assumption on the observation model is as follows:
\begin{equation}
  y_{m,n}=f_m(\bm{x}_{m,n})+\epsilon_{m,n}.
	\label{eq:obs2}
\end{equation}
The residual error $\epsilon_{m,n}\in\mathbb{R}$ is assumed to be a white Gaussian noise that follows $\mathcal{N}(0,\sigma_m^2)$.
The function $f_m$ is tied to each node and assumed to follow MOGP.
In MOGP, correlated $M$ functions $\bm{f}(\cdot)=[f_1(\cdot), f_2(\cdot), \ldots, f_M(\cdot)]^\mathrm{T}$ are considered simultaneously.
The $M$ functions follow a Gaussian process 
\begin{equation}
  \bm{f}(\bm{x})\sim\mathcal{GP}(\bm{0},\bm{k}_\mathcal{M}(\bm{x},\bm{x}')),
\end{equation}
where $\bm{k}_\mathcal{M}$ is the multi-output covariance defined as 
\begin{equation}
	\bm{k}_\mathcal{M}(\bm{x},\bm{x}')=\begin{bmatrix}
		k_{11}(\bm{x},\bm{x}') & k_{12}(\bm{x},\bm{x}') & \cdots & k_{1M}(\bm{x},\bm{x}') \\ 
		\vdots & \vdots & \ddots & \vdots \\
		k_{M1}(\bm{x},\bm{x}') & k_{M2}(\bm{x},\bm{x}') & \cdots & k_{MM}(\bm{x},\bm{x}')
	\end{bmatrix}
\end{equation}
The covariance $k_{m,m'}(\bm{x},\bm{x}')$ represents the correlation between $f_m(\bm{x})$ and $f_{m'}(\bm{x}')$.

The key idea for graph-structured data is to design the covariance $k_{mm'}(\cdot,\cdot)$. 
This covariance should capture two types of correlations: the correlation between vertices of the graph and the correlation between associated data.
By discussing the treatment of graph-structured data from the general formulation of MOGP, 
it is possible to capture a variety of correlations that have not been handled in existing studies.
Specific examples are given in Sect.~\ref{sec:kernel} and the relationship to existing studies is discussed in Sect.~\ref{sec:relation}.

The distribution of $\bm{Y}=[\bm{Y}_1^\mathrm{T},\ldots,\bm{Y}_M^\mathrm{T}]^\mathrm{T}\in\mathbb{R}^{N}$ is given by 
\begin{equation}
  p(\bm{Y})=\mathcal{N}\left(\bm{0}_N,\bm{K}_\mathcal{M}(\bm{X}) + \bm{\Sigma}\right),
\end{equation}
where $\bm{X}=[\bm{X}_1^\mathrm{T},\ldots,\bm{X}_M^\mathrm{T}]^\mathrm{T}\in\mathbb{R}^{N\times D}$, $\bm{\Sigma}=\mathrm{diag}[\sigma_1^2\bm{I}_{N_1},\ldots,\sigma_M^2\bm{I}_{N_M}]\in\mathbb{R}^{N\times N}$ and 
\begin{align}
  &\bm{K}_\mathcal{M}(\bm{X})=\nonumber \\
  &\begin{bmatrix}
    k_{11}(\bm{X}_1)&k_{12}(\bm{X}_1,\bm{X}_2)&\cdots&k_{1M}(\bm{X}_1,\bm{X}_M)\\
    k_{21}(\bm{X}_2,\bm{X}_1)&k_{22}(\bm{X}_2)&\cdots&k_{2M}(\bm{X}_2,\bm{X}_M)\\
    \vdots&\vdots&\ddots&\vdots\\
    k_{M1}(\bm{X}_M,\bm{X}_1)&k_{M2}(\bm{X}_M,\bm{X}_2)&\cdots&k_{MM}(\bm{X}_M)
  \end{bmatrix}.
  \label{eq:kmogp}
\end{align}

One example of the MOGP regression is the prediction of the new output values corresponding to each vertex.
Another context is introduced in Sect. \ref{sec:semi}.
The target values are $\bm{Y}^\ast=[\bm{Y}_1^\ast,\ldots,\bm{Y}_M^\ast]^{\mathrm{T}}\in\mathbb{R}^T$ 
where $\bm{Y}^\ast_m$ is the values from $f_m$ at $T_m$ test points $\bm{X}^\ast_m$ 
and we re-define $T=\sum_{m=1}^M T_m$, 
and we set $\bm{X}^\ast = [\bm{X}_1^\ast,\ldots,\bm{X}_M^\ast]^{\mathrm{T}}\in\mathbb{R}^{T\times D}$.
The predictive distribution of $\bm{Y}^\ast$ is, as is the case of SOGP regression, given by 
\begin{equation}
  p(\bm{Y}^\ast|\bm{X}^\ast,\mathcal{D}_\mathcal{M}) = \mathcal{N}\left(\bm{\mu}_\mathrm{MO}(\bm{X}^\ast),\bm{\Sigma}_\mathrm{MO}(\bm{X}^\ast)\right),
\end{equation}
where $\bm{\mu}_\mathrm{MO}(\bm{X}^\ast)$ and $\bm{\Sigma}_\mathrm{MO}(\bm{X}^\ast)$ are 
\begin{align}
  \bm{\mu}_\mathrm{MO}(\bm{X}^\ast) &= \bm{K}_\mathcal{M}(\bm{X}^\ast,\bm{X})(\bm{K}_\mathcal{M}(\bm{X})+\bm{\Sigma})^{-1} \bm{Y},
  \label{eq:mufull2}\\
  \bm{\Sigma}_\mathrm{MO}(\bm{X}^\ast) &= \bm{\Sigma}^\ast+\bm{K}_\mathcal{M}(\bm{X}^\ast) \nonumber\\
  &- \bm{K}_\mathcal{M}(\bm{X}^\ast,\bm{X}) (\bm{K}_\mathcal{M}(\bm{X})+\bm{\Sigma})^{-1} \bm{K}_\mathcal{M}(\bm{X},\bm{X}^\ast),
  \label{eq:sigma2full2}
\end{align}
where $\bm{\Sigma}^\ast = \mathrm{diag}[\sigma_1^2\bm{I}_{T_1},\ldots,\sigma_M^2\bm{I}_{T_M}]$.

Hyperparameter training is performed by maximizing the log-marginal likelihood of the joint distribution of all outputs:
\begin{align}
  \log{p(\bm{Y}|\bm{X},\bm{\Theta})} &\propto -\frac{1}{2}\Big(\bm{Y}^{\mathrm{T}}(\bm{K}_\mathcal{M}(\bm{X})+\bm{\Sigma})^{-1}\bm{Y} \nonumber\\
  &\quad + \log{\mathrm{det} \left[\bm{K}_\mathcal{M}(\bm{X})+\bm{\Sigma}\right]}\Big).
  \label{eq:ml2}
\end{align}

The difference from SOGP is that the covariance matrix is a block matrix consisting of multiple covariance functions.
In SOGP, $k_{mm'}(\cdot,\cdot)=\delta_{mm'}k(\cdot,\cdot)$ and $\sigma_m^2=\sigma^2$, 
i.e., the common covariance and noise variance are used and the covariance between different outputs is not considered.
The MOGP model can leverage certain higher-order correlations, such as graph structure, not just between data.

As shown in Table~\ref{tab:contributions}, 
formulating graph-structured data regression from the definition of MOGP, as introduced in this section, 
allows for expanded application to a wider range of data configurations, sophisticated correlation extractions, and inference scenarios, 
which are not covered by the conventional MOGP-based methods \cite{Venkitaraman,Zhi}.
Specific extensions are given in the following subsections and the performance of the proposed method for these problems is evaluated in Sect.~\ref{sec:experiments}.

\subsection{Extensions of Acceptable Scenarios and Data Configurations}
\label{sec:input}
The MOGP formulation in Sect.~\ref{sec:formulation} can be applied to a wide range of estimation problems with graph-structured data 
because it encompasses both of the following data configurations and scenarios.

MOGP can handle both heterotopic data and isotopic data.
The \emph{heterotopic data} is the case where each node has node-specific input data, 
i.e., $\bm{X}=[\bm{X}_1^\mathrm{T},\ldots,\bm{X}_M^\mathrm{T}]^\mathrm{T}$ 
and the covariance matrix is as shown in \eqref{eq:kmogp}.
In the case of the \emph{isotopic data}, $\bm{X}_m=\bar{\bm{X}}$ for all $m$ 
and then the covariance matrix has the following form:
\begin{equation}
  \bm{K}_\mathcal{M}(\bm{X})=\begin{bmatrix}
    k_{11}(\bar{\bm{X}})&k_{12}(\bar{\bm{X}})&\cdots&k_{1M}(\bar{\bm{X}})\\
    k_{21}(\bar{\bm{X}})&k_{22}(\bar{\bm{X}})&\cdots&k_{2M}(\bar{\bm{X}})\\
    \vdots&\vdots&\ddots&\vdots\\
    k_{M1}(\bar{\bm{X}})&k_{M2}(\bar{\bm{X}})&\cdots&k_{MM}(\bar{\bm{X}})
  \end{bmatrix}.
\end{equation}
The isotopic data configuration is assumed in the existing studies \cite{Venkitaraman,Zhi} and is related to separable kernels described in later subsection.

There are two scenarios in MOGP: symmetric and assymmetric.
The \emph{symmetric scenario} is popular in the existing studies \cite{Borchani,Venkitaraman,Zhi} 
and is the case where the number of training points is the same for all nodes, 
i.e., $N_m=N/M$ for all $m$.
The \emph{asymmetric scenario} can happen for the heterotopic data and is the case where the number of training points is different for each node.
The covariance formulation given in \eqref{eq:kmogp} is for the asymmetric scenario and includes the symmetric scenario as a special case.

\subsection{Adoption of Sum of Separable Kernels}
\label{sec:kernel}
The key point of MOGP regression for graph-structured data is to design the covariance $k_{mm'}(\cdot,\cdot)$ 
as the correlation that includes both between vertices of the graph and between associated data.
It is possible to make sophisticated kernel selections by considering the MOGP formulation.

One possible way to design is to separate the covariance into the product of the covariance between vertices and the covariance between data.
Such a kernel is called \emph{separable kernel} \cite{MOGP} and formulated as 
\begin{equation}
  k_{mm'}(\bm{x},\bm{x}') = k(\bm{x},\bm{x}')k_{\mathrm{G}}(m,m').
  \label{eq:separable}
\end{equation}
For graph-structured data, 
the correlation between vertices $m$ and $m'$ is decoupled from the total correlations.
Covariance functions known as \emph{graph kernel} can be employed as $k_{\mathrm{G}}(m,m')$.
On the other hand, 
the covariance function $k(\bm{x},\bm{x}')$ is a function for input space alone
and can be chosen from various types of covariance functions for SOGP, such as SE, Mat\'{e}rn, and periodic kernels.

If the data configuration is isotopic, 
the covariance matrix $\bm{K}_\mathcal{M}(\bm{X})$ using the separable kernel can be simplified to 
\begin{equation}
  \bm{K}_\mathcal{M}(\bm{X}) = \bm{K}_{\mathrm{G}}\otimes\bm{K}(\bar{\bm{X}}),
  \label{eq:covseparable}
\end{equation}
where 
\begin{equation}
  \bm{K}_{\mathrm{G}} = \begin{bmatrix}
    k_{\mathrm{G}}(1)&\cdots&k_{\mathrm{G}}(1,M)\\
    \vdots&\ddots&\vdots\\
    k_{\mathrm{G}}(M,1)&\cdots&k_{\mathrm{G}}(M)
  \end{bmatrix}.
\end{equation}
The use of separable kernel is strongly related to the conventional regression methods on graph \cite{Venkitaraman,Zhi}, 
as discussed later in Sect.~\ref{sec:relation}.

On the basis of the fact that the sum of kernels is also a kernel \cite{GPML}, 
we can also consider exploiting the sum of separable (SoS) kernels \cite{MOGP}: 
\begin{equation}
  k_{mm'}(\bm{x},\bm{x}') = \sum_{q=1}^Q k_q(\bm{x},\bm{x}')k_{\mathrm{G},q}(m,m').
  \label{eq:sumofseparable}
\end{equation}
This formulation corresponds to expressing data correlation with $Q$ latent functions.
By preparing appropriate types of kernels according to the characteristics of the data, 
e.g., periodicity or trend, 
and to how they work on the graph, 
it is possible to capture nonlinearities more flexibly and improve the performance of the regression.
For the isotopic data, the covariance becomes 
\begin{equation}
  \bm{K}_\mathcal{M}(\bm{X}) = \sum_{q=1}^Q \bm{K}_{\mathrm{G},q}\otimes\bm{K}_q(\bar{\bm{X}}),
\end{equation}
where $\bm{K}_{\mathrm{G},q}$ and $\bm{K}_q(\bar{\bm{X}})$ are the matrix forms of $k_{\mathrm{G},q}(m,m')$ and $k_q(\bm{x},\bm{x}')$, respectively.
The effect of this extension will be confirmed by experiments in Sect.~\ref{sec:experiments}.

\subsection{Adoption of Kernels with Convolution}
\label{sec:conv}
In this subsection, 
we propose a novel kernel design for graph-structured data on the basis of convolution of kernels.

The above separable constructions are simple but it is known that there are limitations in some applications in the context of MOGP \cite{MOGP,Liu}.
This is due to the fact that the separable structure is derived from 
the assumption that the latent functions corresponding to each vertex are independent of each other 
but share the same data covariance $k(\cdot,\cdot)$.
One extention is to consider the convolution of these latent functions.
Process convolution (PC) \cite{PC} is one of the methods to consider the convolution of latent functions.
It is possible to overcome explicit sharing of the data kernel $k(\cdot,\cdot)$ 
and to own hyperparameters related to data, such as length scale, differently for each vertex.
The PC is highly flexible, allowing a wide choice of function shapes, 
but assuming the smooth kernels and a Gaussian form function, the kernel is given by 
\begin{align}
  k_{mm'}(\bm{x},\bm{x}') &= \frac{v^2 s_m s_{m'}}{(2\pi)^{D/2}|\bm{P}|^{1/2}} \nonumber\\
  &\quad \cdot \exp\left(-\frac{1}{2}(\bm{x}-\bm{x}')^\mathrm{T}\bm{P}^{-1}(\bm{x}-\bm{x}')\right),
  \label{eq:processconv}
\end{align}
where $\bm{P}=\bm{P}_m^{-1}+\bm{P}_{m'}^{-1}+\bm{\Lambda}^{-1}\in\mathbb{R}^{D\times D}$, 
and where $v,s_m, s_{m'}\in\mathbb{R}$ and $\bm{P}_m,\bm{P}_{m'},\bm{\Lambda}\in\mathbb{R}^{D\times D}$ 
are hyperparameters.

The terms $s_m s_{m'}$ and $\bm{P}_m^{-1}+\bm{P}_{m'}^{-1}$ are related to the correlation between vertices $m$ and $m'$.
Therefore, for graph-structured data, it is expected to be useful to include information on correlations between vertices derived from the graph structure in these terms.
We then propose to use a graph kernel as $s_m s_{m'}$ 
and to determine $\bm{P}_m^{-1}+\bm{P}_{m'}^{-1}$ by another graph kernel.
We name the PC designed with graph structure information as \emph{graph PC}.

The component $s_m s_{m'}$ is scalar so that the graph kernel itself can be used as $s_m s_{m'}$.
That is, we can choose 
\begin{equation}
  s_m s_{m'} = k_{\mathrm{G},1}(m,m').
  \label{eq:ss}
\end{equation}

On the other hand, there are wide range of choices for the design of $\bm{P}_m^{-1}+\bm{P}_{m'}^{-1}$ 
as long as symmetry and positivity as a kernel function are not compromised.
One of the simplest choices is to use a graph kernel $k_{\mathrm{G}}(m,m')$ as 
\begin{equation}
  \bm{P}_m^{-1}+\bm{P}_{m'}^{-1} = k_{\mathrm{G},2}(m,m')^{-1}\bm{I}_D.
  \label{eq:pp}
\end{equation}
Alternatively, parameters $p_m$ and $p_{m'}$ containing vertex-specific information, such as the degree of each vertex, 
can be prepared, and the formulation is given by 
\begin{equation}
  \bm{P}_m^{-1}+\bm{P}_{m'}^{-1} = p_m^{-1}\bm{I}_D + p_{m'}^{-1}\bm{I}_D.
\end{equation}
Regardless of the above use cases, 
a wide range of applications can be considered by incorporating graph information into the hyperparameters.

\subsection{Extension of Inference Scenarios}
\label{sec:semi}
The generalized formulation of MOGP for graph-structured data can also extend inference scenarios.
There are several examples: 
The proposed formulation can be extended to multi-class classification problems \cite{Ma} as well as regression.
For the case where the outputs on graph have different levels of fidelity, 
known as hierarchical multi-fidelity scenario \cite{Liu}, 
the proposed formulation can design kernels to capture the difference \cite{Kennedy,Myers,Leen}.
It is also possible to consider the effective use of graph information 
for the multiple-output version of the Gaussian process latent variable model \cite{Hu,Muk}, which is an unsupervised learning method.

Specifically, we introduce in this subsection the case 
where the relevant vertices are different for training and test phases, respectively.
A typical example is missing data estimation in sensor networks, 
where we want to estimate missing values of some sensor nodes 
by using previous observations of the sensors and all observations obtained for the other sensors.
In this case, all nodes including the defective sensor nodes are used in the training phase, 
and predictions are made only for the defective nodes.
The conventional methods \cite{Venkitaraman,Zhi} do not cover such a problem 
because they assume that vertices are the same for training and test phases, 
which is specifically discussed in Sect.~\ref{sec:relation}.
In other words, the conventional methods require reformulating such a problem 
into a simpler one, where the prediction of missing values is performed using only the past observations of the defective nodes themselves as output.
Considering the proposed formulation on the basis of the definition of MOGP, 
it is possible to predict missing values using all of the available data.

Consider the case where a graph $\mathcal{G}=(\mathcal{V},\mathcal{E})$ $(|\mathcal{V}|=M)$ 
and dataset $\mathcal{D}=\{\mathcal{D}_1,\ldots,\mathcal{D}_M\}$ are given, 
and the target values are $\tilde{\bm{Y}}^\ast = [\bm{Y}_{v_1}^{\ast\mathrm{T}},\ldots,\bm{Y}_{v_O}^{\ast\mathrm{T}}]^{\mathrm{T}}$ 
corresponding to the test inputs $\tilde{\bm{X}}^\ast = [\bm{X}_{v_1}^{\ast\mathrm{T}},\ldots,\bm{X}_{v_O}^{\ast\mathrm{T}}]^{\mathrm{T}}$ 
for some subset of $O(<M)$ nodes, $\tilde{\mathcal{V}}=\{v_1,\ldots,v_O\}\subset\mathcal{V}$.
For each vertex $v_o\in\tilde{\mathcal{V}}$, there are $T_{v_o}$ test points and $\tilde{T}=\sum_{o=1}^O T_{v_o}$.
In this case, the proposed MOGP formulation allows regression on induced subgraph $\tilde{\mathcal{G}}=(\tilde{\mathcal{V}},\tilde{\mathcal{E}})\subset\mathcal{G}$ 
using the training data on $\mathcal{G}$. 
The joint distribution of $\bm{Y}$ and $\tilde{\bm{Y}}^\ast$ is given by 
\begin{align}
  &p(\begin{bmatrix}\bm{Y}\\
  \tilde{\bm{Y}}^\ast
  \end{bmatrix}) \nonumber\\
  &= \mathcal{N}\left(
  \bm{0},
  \begin{bmatrix}
    \bm{K}_\mathcal{M}(\bm{X})&\bm{K}_\mathcal{MO}(\bm{X},\tilde{\bm{X}}^\ast)\\
    \bm{K}_\mathcal{MO}(\bm{X},\tilde{\bm{X}}^\ast)^\mathrm{T}&\bm{K}_\mathcal{O}(\tilde{\bm{X}}^\ast)
  \end{bmatrix} + \sigma^2\bm{I}_{N+\tilde{T}}\right), 
\end{align}
where 
\begin{align}
  &\bm{K}_\mathcal{MO}(\bm{X},\tilde{\bm{X}}^\ast) \nonumber\\
  &= \begin{bmatrix}
    k_{1 v_1}(\bm{X}_1,\tilde{\bm{X}_{v_1}^\ast})&\cdots&k_{1 v_O}(\bm{X}_1,\tilde{\bm{X}_{v_O}^\ast})\\
    \vdots&\ddots&\vdots\\
    k_{M v_1}(\bm{X}_M,\tilde{\bm{X}_{v_1}^\ast})&\cdots&k_{M v_O}(\bm{X}_M,\tilde{\bm{X}_{v_O}^\ast})
  \end{bmatrix},
\end{align}
and 
\begin{align}
  &\bm{K}_\mathcal{O}(\tilde{\bm{X}}^\ast) \nonumber\\
  &= \begin{bmatrix}
    k_{v_1 v_1}(\tilde{\bm{X}_{v_1}^\ast})&\cdots&k_{v_1 v_O}(\tilde{\bm{X}_{v_1}^\ast},\tilde{\bm{X}_{v_O}^\ast})\\
    \vdots&\ddots&\vdots\\
    k_{v_O v_1}(\tilde{\bm{X}_{v_O}^\ast},\tilde{\bm{X}_{v_1}^\ast})&\cdots&k_{v_O v_O}(\tilde{\bm{X}_{v_O}^\ast})
  \end{bmatrix}.
\end{align}
The predictive distribution of $\tilde{\bm{Y}}^\ast$ is given by 
\begin{equation}
  p(\tilde{\bm{Y}}^\ast|\tilde{\bm{X}}^\ast,\mathcal{D}) = \mathcal{N}\left(\bm{\mu}_\mathrm{IS}(\tilde{\bm{X}}^\ast),\bm{\Sigma}_\mathrm{IS}(\tilde{\bm{X}}^\ast)\right),
\end{equation}
where 
\begin{equation}
  \bm{\mu}_\mathrm{IS}(\tilde{\bm{X}}^\ast) = \bm{K}_\mathcal{MO}(\bm{X},\tilde{\bm{X}}^\ast)^\mathrm{T}(\bm{K}_\mathcal{M}(\bm{X})+\bm{\Sigma})^{-1}\tilde{\bm{Y}}^\ast,
\end{equation}
and 
\begin{align}
  &\bm{\Sigma}_\mathrm{IS}(\tilde{\bm{X}}^\ast) = \bm{K}_\mathcal{O}(\tilde{\bm{X}}^\ast) \nonumber\\
  &- \bm{K}_\mathcal{MO}(\bm{X},\tilde{\bm{X}}^\ast)^\mathrm{T}(\bm{K}_\mathcal{M}(\bm{X})+\bm{\Sigma})^{-1}\bm{K}_\mathcal{MO}(\bm{X},\tilde{\bm{X}}^\ast).
\end{align}

The correlation among all nodes are included in the training covariance matrix $\bm{K}_\mathcal{M}(\bm{X})$.
The information on the induced subgraph structure for prediction is appeared as $k_{v v_o}(\bm{X}_v,\tilde{\bm{X}}_{v_o}^\ast)$ in the covariance matrix.
This means that the model can incorporate information on both the original graph and the induced subgraph.
That is, all the available data are incorporated in the training phase 
and the model can incorporate information on both the original graph and the induced subgraphs.

\subsection{Relation to Existing Gaussian Processes on Graph}
\label{sec:relation}
SOGP regression is included as a special case of the proposed MOGP formulation.
If one adopts the separable kernel in eq.~\eqref{eq:separable} and sets $k_{\mathrm{G}}(m,m')=\delta_{mm'}$, 
the proposed MOGP formulation shrinks to SOGP regression.
For the SOGP regression, there are several studies that have proposed graph kernels \cite{Smola}.
The kernel $k(\cdot,\cdot)$ in eq.~\eqref{eq:full} is, in this case, characterized only by the graph structure.
On the other hand, the kernel $k(\cdot,\cdot)$ in eq.~\eqref{eq:kmogp} of the proposed MOGP formulation can incorporate both information of the graph structure and data.
This leads to extend the expressive potential of the model.

MOGP-based regression methods for graph-structured data proposed in \cite{Venkitaraman,Zhi} are also included as special cases of the proposed formulation.
These are based on the following graph signal model:
\begin{equation}
  \bm{y}_n=\bm{C}\bm{f}(\bm{x}_n)+\bm{\epsilon}_n,
\end{equation}
where $\bm{y}_n\in\mathbb{R}^M$ and $\bm{x}_n\in\mathbb{R}^D$ are $n$th output and input data $(n=1,2,\ldots,N)$ on $M$ vertices, respectively, 
$\bm{C}\in\mathbb{R}^{M\times M}$ is a graph filtering matrix, 
$\bm{f}:\mathbb{R}^D\to\mathbb{R}^M$ is a function the elements of which are assumed to be independent GPs with a common kernel function, 
and $\bm{\epsilon}_n$ is the additive noise vector.
This model assumes isotopic data $\bar{\bm{X}}=\{\bm{x}_1,\ldots,\bm{x}_N\}$ and symmetric scenario.
According to the model, the covariance matrix in terms of the stacked vector $\bm{F}=\mathrm{vec}([\bm{f}(\bm{x}_1),\ldots,\bm{f}(\bm{x}_N)]^\mathrm{T})$ 
corresponds to $\mathrm{Cov}(\bm{F},\bm{F})=\bm{K}_\mathcal{M}(\bm{X})$ and can be summarized as 
\begin{equation}
  \bm{K}_\mathcal{M}(\bm{X}) = \bm{CC}^\mathrm{T}\otimes\bm{K}(\bar{\bm{X}}).
\end{equation}
This is included in the covariance formulation of separable kernel \eqref{eq:covseparable} where $\bm{K}_\mathrm{G}=\bm{CC}^\mathrm{T}$, 
as mentioned in the papers \cite{Venkitaraman,Zhi}.
The designs of the graph filter $\bm{C}$ are specifically discussed in \cite{Venkitaraman,Zhi}.
The global filtering kernel is used for \cite{Venkitaraman} 
and the polynomial kernel is for \cite{Zhi}.
The specific formulations of these kernels are summarized in Table~\ref{tab:graphkernels} in Sect.~\ref{sec:experiments}.

In summary, the methods correspond to one case of the MOGP-based models in Sect.~\ref{sec:formulation} 
where the data configuration is isotopic and in symmetric scenario, 
and where one of the separable kernels are selected.

\subsection{Computational Complexity and Number of Hyperparameters}
\label{sec:complexity}
In this subsection, we discuss the computational complexity of MOGP regression.

The inverse computation of the covariance matrix dominates the computational complexity of inference in Gaussian process regression. 
However, as long as the same amount of data is used, 
SOGP and MOGP have the same computational complexity because the size of the covariance matrix remains the same.
Similarly, the choice of model in MOGP (ICM or the aforementioned kernel) does not change the computational complexity of the inverse.
The time complexity of the inverse computation is $O(N^3)$ because of the $N\times N$ covariance matrix.

The number of hyperparameters affects the execution time in practice in the hyperparameter training phase.
Since the detailed complexity depends on the stopping conditions and algorithms, 
we discuss here the number of hyperparameters that need to be optimized.
MOGP for graph-structured data generally requires additional hyperparameters 
in addition to ones required in SOGP (e.g., $v$, $\ell$ of SE kernel, and noise variance $\sigma^2$).
We set $N_h (>0)$ as the number of hyperparameters for $\bm{K}$ and $N_g (>0)$ as that for $\bm{K}_{\mathrm{G}}$.
The number $N_g$ is a few in general, which does not depends on graph topology; 
Table~\ref{tab:graphkernels} in Sect.~\ref{sec:experiments} summarizes representative graph kernels for MOGP and shows that $N_g=0$ or $1$ in the most cases.
On the other hand, the number of hyperparameters in ICM (rank-$1$ case) is $N_g=2M$ that depends on the number of vertices.
Although the flexibility of the model is high, there is concern that, for large graphs, the optimization can be time-consuming and model fitting may be difficult.
Table~\ref{tab:hyp} shows that the number of hyperparameters in SOGP, ICM, and MOGP with separable kernel or SoS kernels \eqref{eq:sumofseparable} and with graph PC.
For graph PC, we show the case when \eqref{eq:ss}, \eqref{eq:pp}, and $\bm{\Lambda}=\ell\bm{I}$ are used.
One of the advantages of considering MOGP for graph-structured data is 
that the model can be represented with a small number of hyperparameters using the valuable prior information of graph structure.
\begin{table}[tb]
  \centering
  \caption{Number of hyperparameters. $N_{h,q}$ is the number for $k_q(\cdot,\cdot)$ and $N_{g,q}$ is that for $k_{\mathrm{G},q}(\cdot,\cdot)$.}
  \label{tab:hyp}
  \begin{tabular}{|c|c|}
    \hline
    Method & Number of hyperparameters \\
    \hline
    SOGP & $N_h$ \\
    ICM & $N_h + 2M$ \\
    MOGP with separable kernel & $N_h + N_g$ \\
    MOGP with SoS kernels & $\sum_{q=1}^Q (N_{h,q} + N_{g,q})$ \\
    MOGP with graph PC & $N_{g,1} + N_{g,2} + 2$ \\
    \hline
  \end{tabular}
\end{table}

\section{Experiments}
\label{sec:experiments}
\subsection{Overview}
In this section, 
specific extensions allowed by the proposed formulation are demonstrated and validated using computer experiments.
The experimental codes written in Python and datasets are available in a GitHub repository\footnote{https://github.com/a-nakai-k/MOGP-for-Graph-Structured-Data}.
All data were preprocessed with the mean set to zero, respectively.
The gradient descent method was used for the hyperparameter optimization.
We set $\sigma^2_1=\sigma^2_2=\ldots=\sigma^2_M=\sigma^2$ in all experiments.

Existing graph kernels used for performance comparisons and the proposed kernel extensions are summarized in Table~\ref{tab:graphkernels}.
These are formulated as the graph kernel $k_{\mathrm{G}}(\cdot,\cdot)$ in the separable kernel \eqref{eq:separable} 
or $k_{\mathrm{G},q}(\cdot,\cdot)$ in the SoS kernels \eqref{eq:sumofseparable}.
In Table~\ref{tab:graphkernels}, we summarize the graph kernels using the following notations:
\begin{equation}
  \bm{K}_{\mathrm{G}} = \bm{B}~\mathrm{or}~\bm{CC}^\mathrm{T}.
\end{equation}
The matrix $\bm{B}\in\mathbb{R}^{M\times M}$ or $\bm{C}\in\mathbb{R}^{M\times M}$ is determined depending on the method.
The matrix $\bm{L}\in\mathbb{R}^{M\times M}$ is a Laplacian matrix of the graph, 
$\tilde{\bm{L}}\in\mathbb{R}^{M\times M}$ is a normalized Laplacian matrix, 
and $\hat{\bm{L}}\in\mathbb{R}^{M\times M}$ means that either $\bm{L}$ or $\tilde{\bm{L}}$ can be used.
The polynomial-$d$ kernel requires additional optimization for satisfying constraints as a kernel \cite{Zhi}.
In addition to these kernels, we employed ICM \cite{ICM}, 
one of the MOGP methods that does not use graph information, 
to demonstrate the benefit of utilizing graph information.
The ICM has the separable kernel structure as in \eqref{eq:separable} and given by 
$\bm{B}=\bm{ww}^\mathrm{T}+\mathrm{diag}[\bm{\kappa}]$, 
where $\bm{w},\bm{\kappa}\in\mathbb{R}^{M\times 1}$ are parameter vectors 
which are determined as the hyperparameters.
Note that we employed the ICM of rank-1 case.
\begin{table}[tb]
  \centering
  \caption{Graph kernels used for performance comparisons and the proposed kernel extensions. Parameters $\alpha$, $\{\beta_i\}$, $\bm{w}$, and $\bm{\kappa}$ are determined by the maximization of the log-likelihood \eqref{eq:ml2}.}
  \label{tab:graphkernels}
  \begin{tabular}{|c|c|c|}
    \hline
    Method & $\bm{K}_{\mathrm{G}}$ & $N_g$ \\
    \hline
    Laplacian \cite{MOGP} & $\bm{B}=\bm{L}^\dagger$ & $0$ \\
    Global filtering \cite{Venkitaraman} & $\bm{C}=(\bm{I}+\alpha\bm{L})^{-1}$ & $1$ \\
    Local averaging \cite{Wilson,Zhi} & $\bm{C}=(\bm{I}+\alpha\bm{D})^{-1}(\bm{I}+\alpha\bm{A})$ & $1$ \\
    Regularized Laplacian \cite{Smola} & $\bm{B}=(\bm{I}+\alpha\tilde{\bm{L}})^{-1}$ & $1$ \\
    Diffusion \cite{Smola} & $\bm{B}=\exp((-\alpha/2)\tilde{\bm{L}})$ & $1$ \\
    $p$-step random walk \cite{Smola} & $\bm{B}=(\alpha\bm{I}-\tilde{\bm{L}})^p$ & $1$ \\
    Cosine \cite{Smola} & $\bm{B}=\cos(\tilde{\bm{L}}\pi/4)$ & $0$ \\
    Graph Mat\'{e}rn-$\nu$ \cite{Boro} & $\bm{B}=((2\nu/\alpha)\bm{I}+\hat{\bm{L}})^{-\nu}$ & $1$ \\
    Polynomial-$d$ \cite{Zhi} & $\bm{C}=\sum_{i=0}^P\frac{\beta_i}{\lambda_{\max}(\bm{L})}\bm{L}^i$ & $P+1$ \\
    ICM \cite{ICM} & $\bm{B}=\bm{ww}^\mathrm{T}+\mathrm{diag}[\bm{\kappa}]$ & $2M$ \\
    \hline
  \end{tabular}
\end{table}

We employed two performance metrics using the target output value $\bm{y}^\ast$, predicted mean $\bm{\mu}^\ast$, and covariance matrix $\bm{\Sigma}^\ast$: 
mean squared error (MSE) and log-likelihood.
The MSE is defined as 
\begin{equation}
  \mathrm{MSE} = \frac{1}{T}\|\bm{y}^\ast-\bm{\mu}^\ast\|_2^2.
\end{equation}
The log-likelihood provides the goodness of the model including the predictive covariance matrix 
and is defined as 
\begin{align}
  \mathrm{Log}\text{-}\mathrm{likelihood} &= -\frac{T}{2}\log(2\pi)-\frac{1}{2}\log\mathrm{det}[\bm{\Sigma}^\ast] \nonumber\\
  &\quad -\frac{1}{2}(\bm{y}^\ast-\bm{\mu}^\ast)^\mathrm{T}\bm{\Sigma}^{\ast -1}(\bm{y}^\ast-\bm{\mu}^\ast).
\end{align}
Lower MSE and higher log-likelihood indicate better performance.

\begin{figure}[tb]
  \centering
  \begin{minipage}[b]{0.49\columnwidth}
      \centering
      \includegraphics[width=0.9\columnwidth]{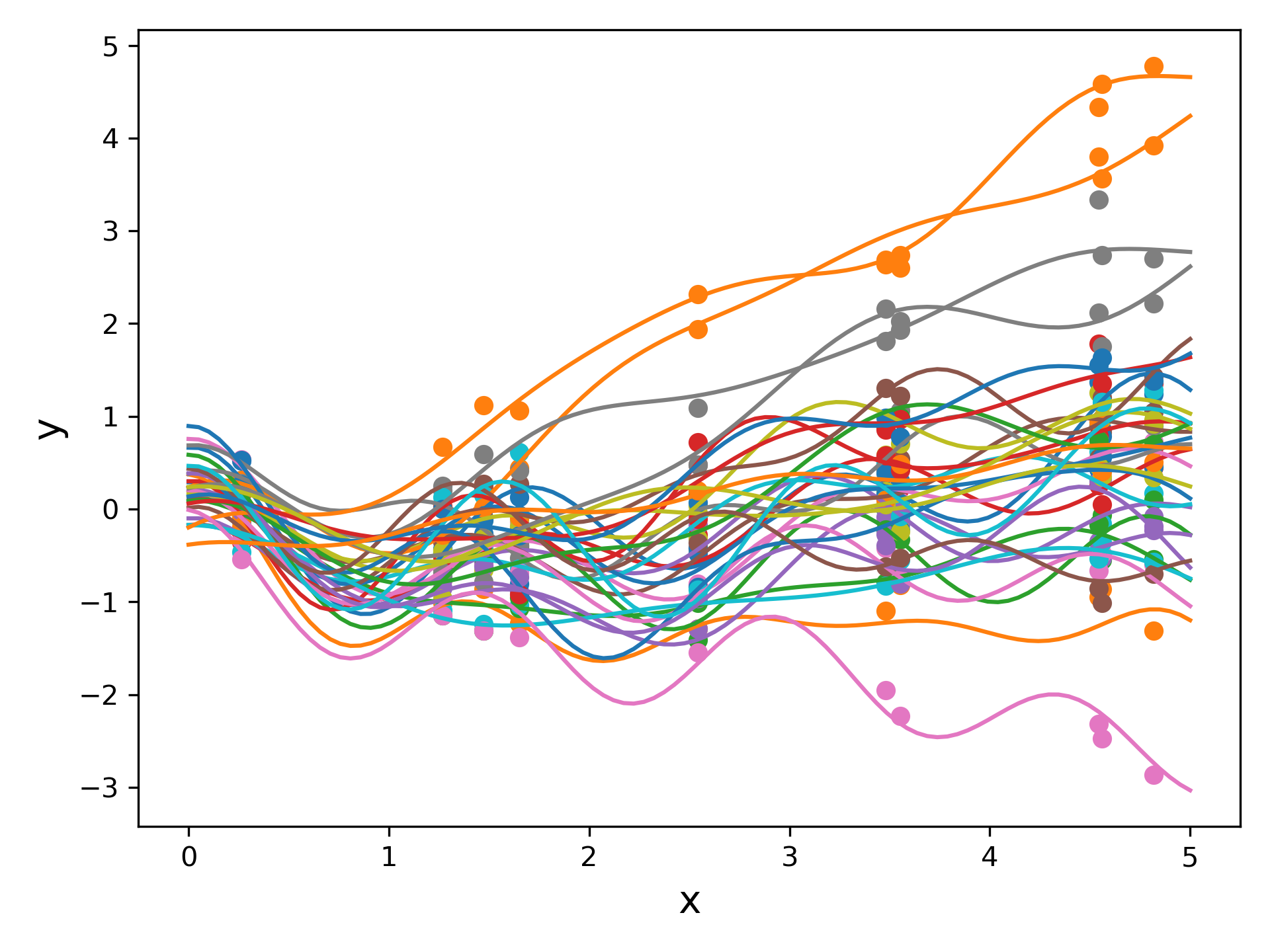}
      \subcaption{$k=6$}
  \end{minipage}
  \begin{minipage}[b]{0.49\columnwidth}
      \centering
      \includegraphics[width=0.9\columnwidth]{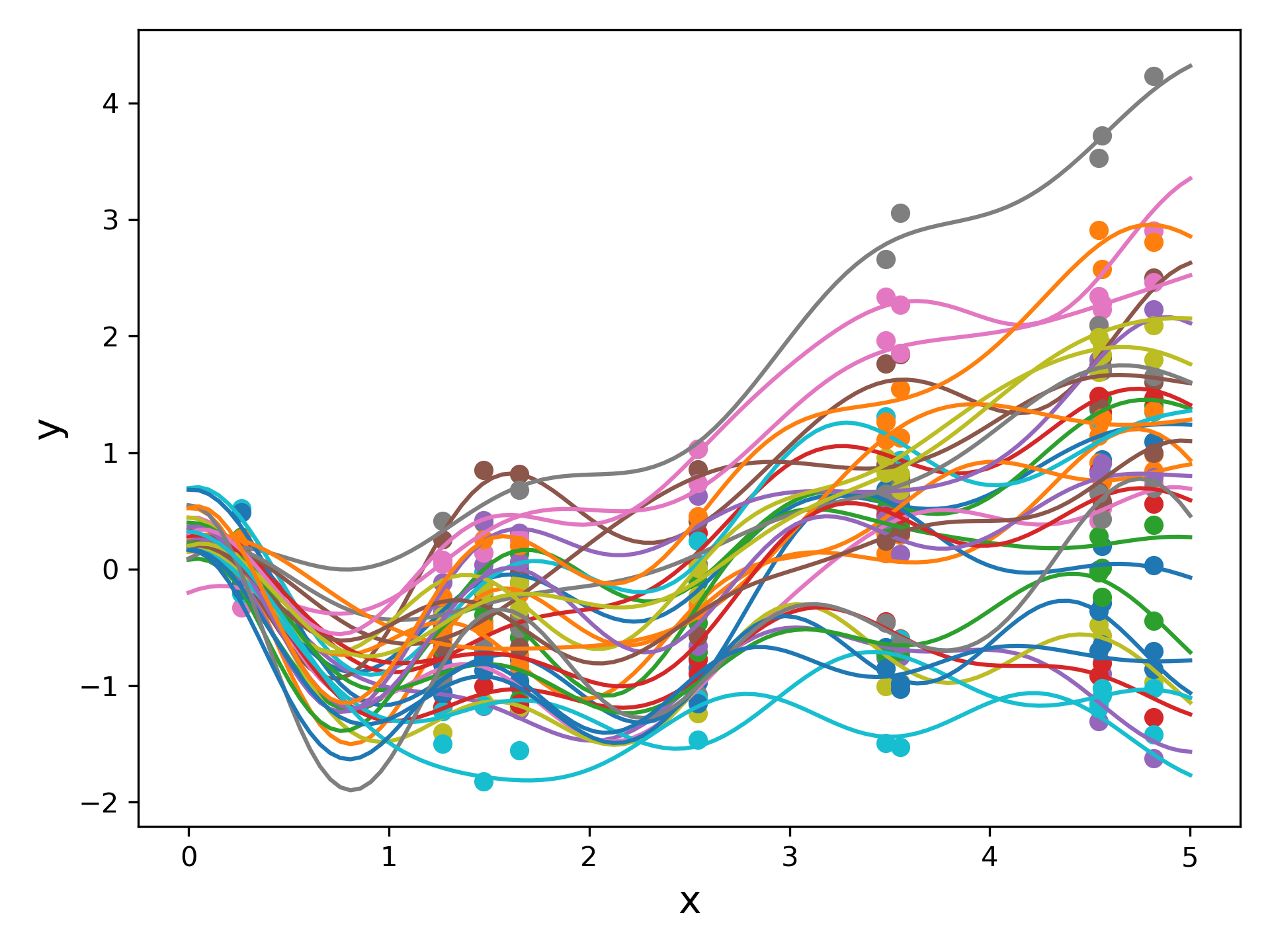}
      \subcaption{$k=12$}
  \end{minipage}\\
  \begin{minipage}[b]{0.49\columnwidth}
      \centering
      \includegraphics[width=0.9\columnwidth]{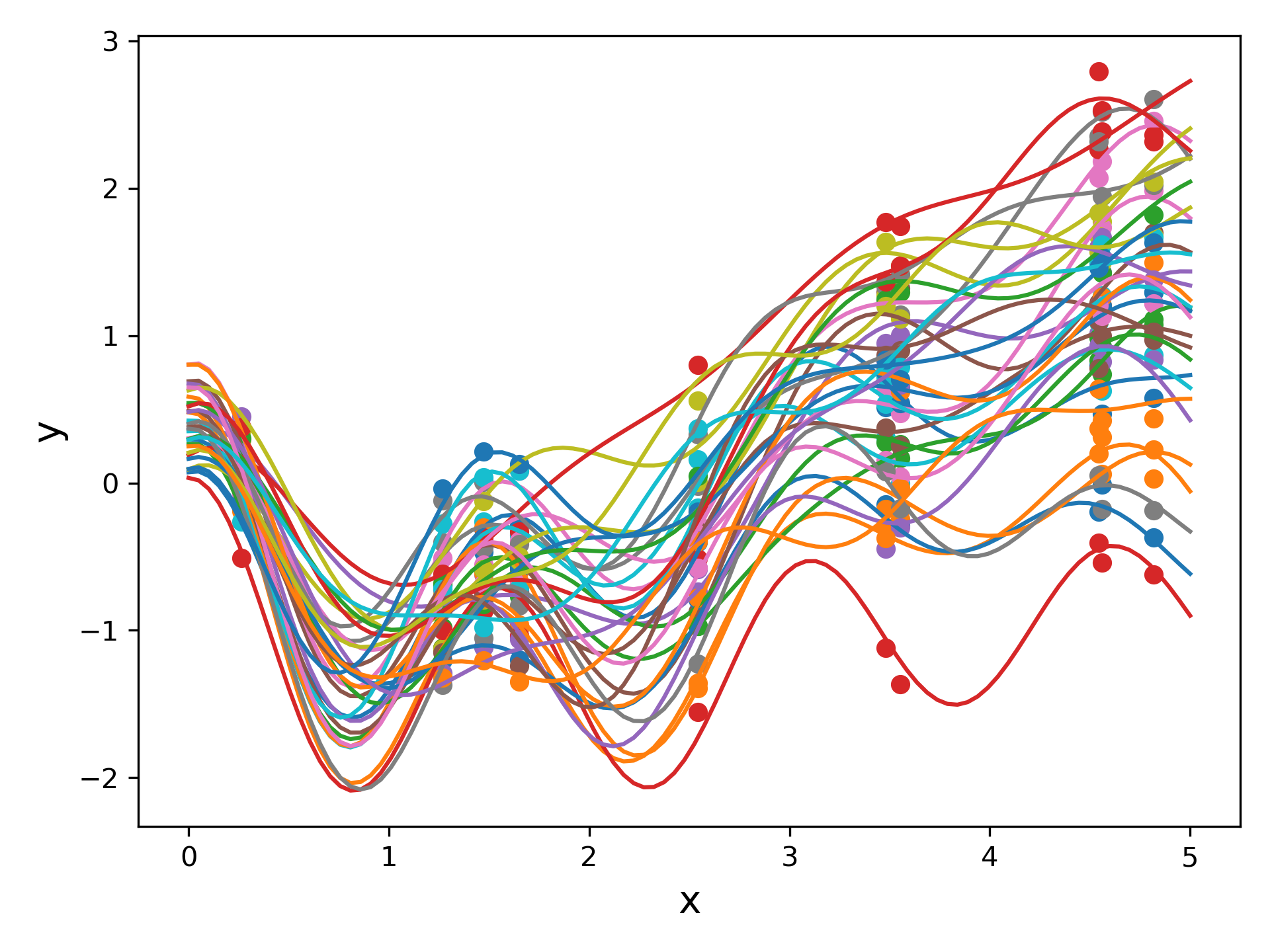}
      \subcaption{$k=18$}
  \end{minipage}
  \begin{minipage}[b]{0.49\columnwidth}
      \centering
      \includegraphics[width=0.9\columnwidth]{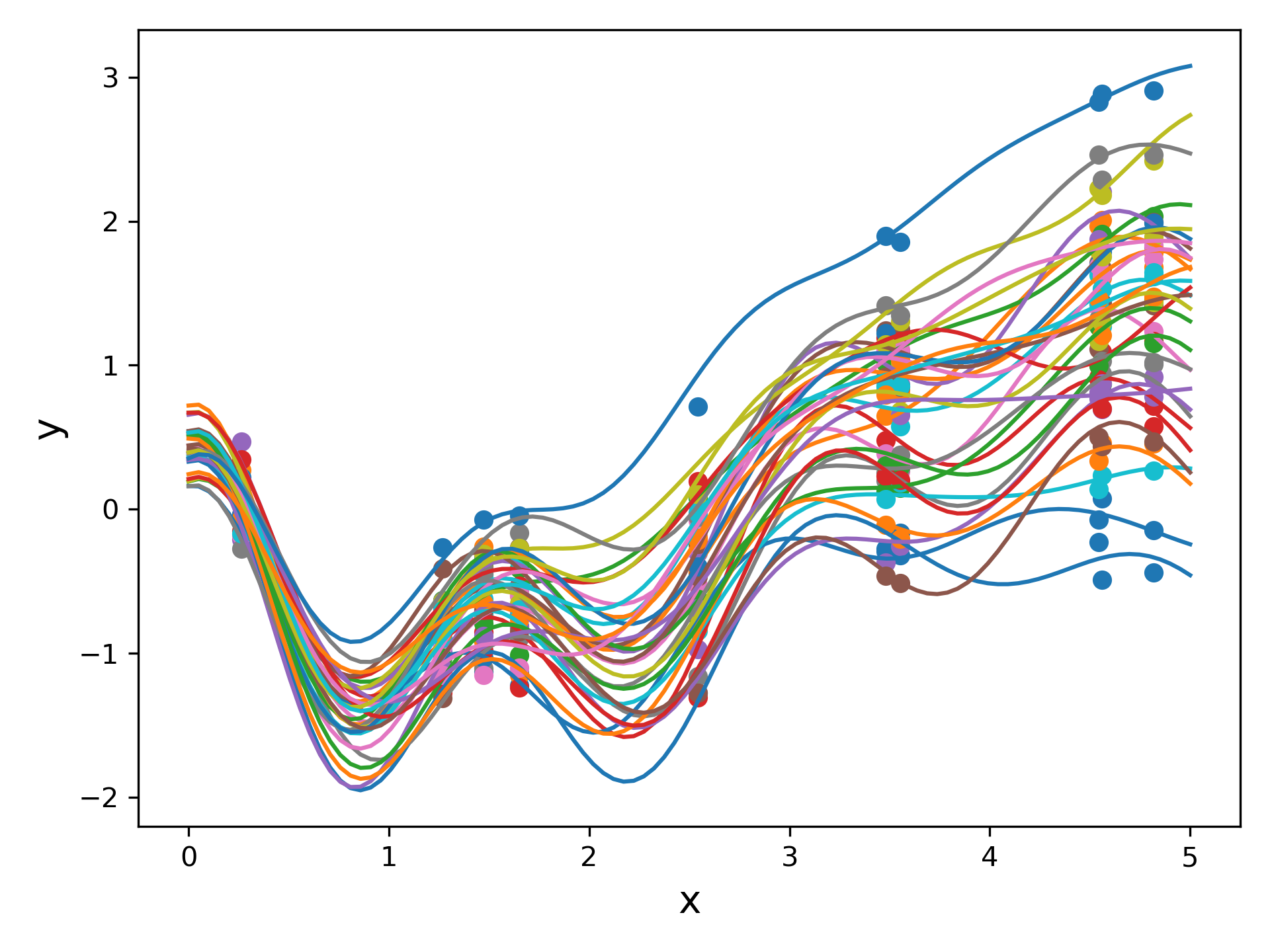}
      \subcaption{$k=24$}
  \end{minipage}
  \caption{Training data for the experiment in Sect.~\ref{sec:synthetic3}. Lines and points are underlying functions and training data for each vertex, respectively.}
  \label{fig:synthetic3}
\end{figure}
\subsection{Verification of Topological Effects with Synthetic Data}
\label{sec:synthetic3}
We first verified with synthetic data the effects of topology on the performance of the MOGP regression for graph-structured data.

We assumed random $k$-regular graphs with $M=32$ vertices and 
the observed data were generated from the following function:
\begin{equation}
  y_{m,n}=\sum_{j\in\mathcal{N}_m\setminus m}(p_j \cos(q_j x_{n})+r_j x_{n}) + \epsilon_{m,n},
\end{equation}
where $\mathcal{N}_m$ is the set of adjacent vertices to $m$, 
$x_n\in[0,5] \ (n=1,2,\ldots,N)$ is the inputs which are common for all vertices, i.e., the isotopic data configuration,
$p_j\in[0,10],q_j\in[5,5],r_j\in[-6,6]$ are uniformly and randomly chosen parameters and fixed for each vertex, 
and $\epsilon_{m,n}$ follows $\mathcal{N}(0,5)$.
Consider the prediction problem introduced in Sect.~\ref{sec:formulation} that the output values corresponding to each vertex.
The topological effects were verified by changing the degree $k$ of the graph, 
which controls the strength of the correlation with adjacent vertices' data.

We prepared degree $k=6,12,18,24$ graphs. 
For each graph, training was performed with randomly chosen $N_m=10$ data points for each vertex $m$.
The underlying functions and the training data are shown in Fig.~\ref{fig:synthetic3}.
Performance of SOGP, ICM, and MOGP for graph (the proposed formulation) was evaluated by $100$ trials of prediction on different randomly chosen $T_m=10$ test points in $[0,5]$.
The SE kernel was used as $k(\cdot,\cdot)$ in all methods.
For the MOGP, we employed the separable kernel structure as in \eqref{eq:separable} and used the local averaging kernel as $k_{\mathrm{G}}(\cdot,\cdot)$.

\begin{figure}[tb]
    \centerline{\includegraphics[width=\columnwidth]{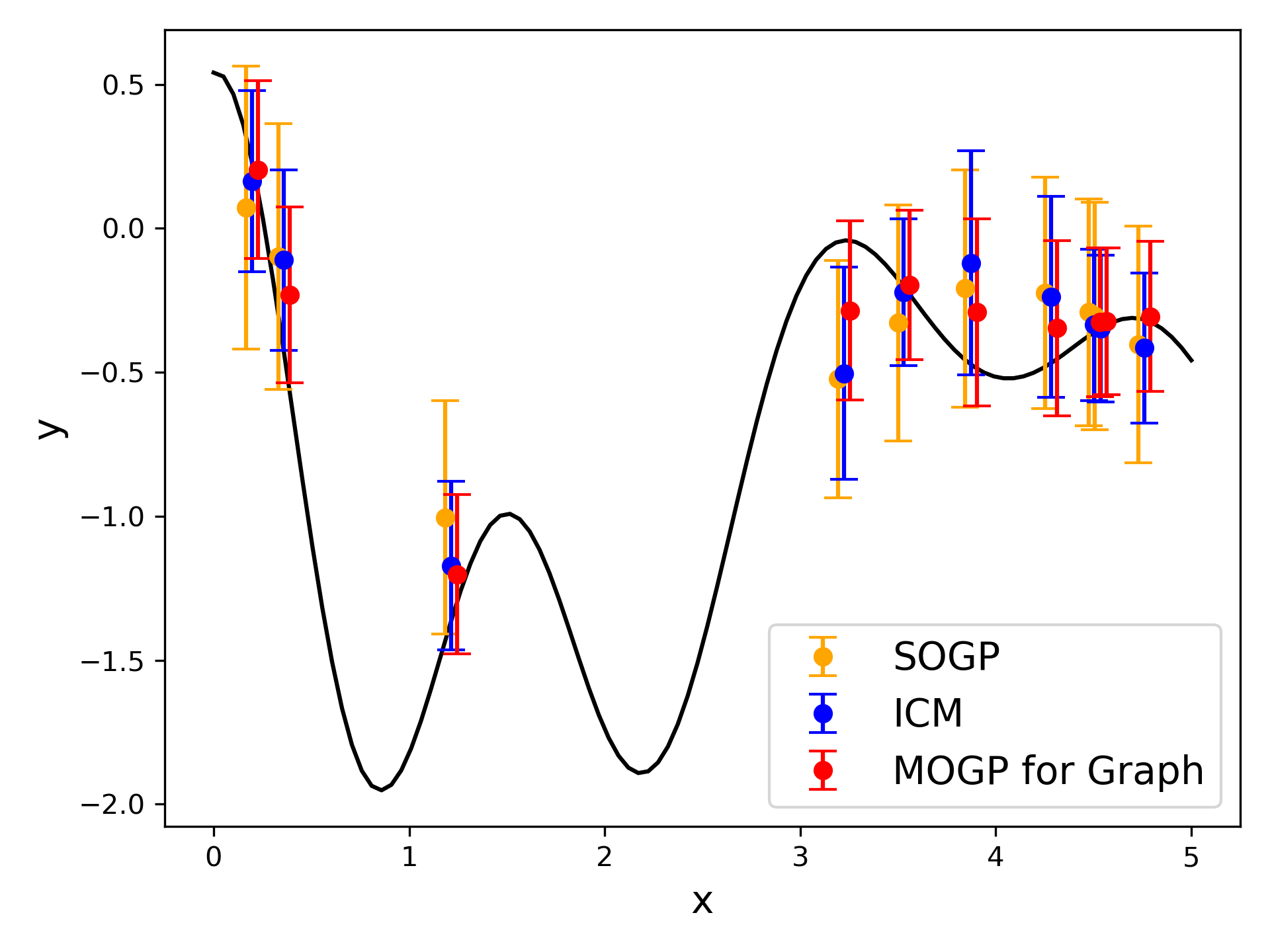}}
    \caption{Predicted mean and 95\% confidence interval of $10$ test points for the case of $k=24$. 
    The $x$ coordinates of the predicted means of SOGP and MOGP for graph are shifted by a small amount for better visibility.}
    \label{fig:synthetic3_plot}
\end{figure}

The mean and standard deviation of each metric are shown in Table~\ref{tab:results_synthetic3}.
The best performance in each case and metric is highlighted in bold.
In most cases, ICM showed better performance than other methods for MSE.
The larger number of hyperparameters of ICM may be the reason of this.
It is noteworthy that the log-likelihood is significantly improved in MOGP for graph-structured data for $k=18$ and $24$.
One example of predicted mean and 95\% confidence interval in the case of $k=24$ is shown in Fig.~\ref{fig:synthetic3_plot}.
By focusing on the right side of the figure, especially near $x=3$, 
we can see that the confidence interval of MOGP for graph-structured data was narrower than those of SOGP and ICM, 
i.e., the confidence level was higher, 
and the underlying function could be captured within the confidence interval.
In other words, the stronger the correlation of data with adjacent vertices, 
the more benefitial the introduction of graph information in MOGP regression for increasing the confidence of the predicted model.
\begin{table}[t]
  \centering
  \caption{Performance of the proposed formulation on the synthetic data in Sect.~\ref{sec:synthetic3}. $\star$ indicates that the method is better than the other methods considering the range of standard deviations.}
  \label{tab:results_synthetic3}
  \begin{tabular}{|c|c|c|c|}
    \hline 
    $k$ & Method & MSE$(\times 10^{-1}) ~ \blacktriangledown$ & Log-likelihood $\blacktriangle$ \\
    \hline
    & SOGP & $1.277\pm 0.0270$ & $-12.42\pm 0.395$ \\
    $6$ & ICM & $1.266\pm 0.0272$ & $-12.95\pm 0.477$ \\
    & MOGP for Graph & $\mathbf{1.261\pm 0.0268}$ & $\mathbf{-11.76\pm 0.385}$ \\
    \hline
    & SOGP & $1.145\pm 0.0434$ & $-10.61\pm 0.774$ \\
    $12$ & ICM & $\mathbf{0.8537\pm 0.0288}\star$ & $\mathbf{-2.465\pm 0.621}\star$ \\
    & MOGP for Graph & $1.113\pm 0.0412$ & $-7.869\pm 0.653$ \\
    \hline
    & SOGP & $1.800\pm 0.0778$ & $-25.43\pm 1.42$ \\
    $18$ & ICM & $\mathbf{0.8846\pm 0.0278}\star$ & $-1.352\pm 0.590$ \\
    & MOGP for Graph & $1.011\pm 0.0324$ & $\mathbf{5.234\pm 0.271}\star$ \\
    \hline
    & SOGP & $1.588\pm 0.0766$ & $-30.21\pm 1.99$ \\
    $24$ & ICM & $\mathbf{0.5060\pm 0.0142}$ & $7.915\pm 0.457$ \\
    & MOGP for Graph & $0.5285\pm 0.0165$ & $\mathbf{15.97\pm 0.219}\star$ \\
    \hline
  \end{tabular}
\end{table}

\begin{figure}[tb]
  \centerline{\includegraphics[width=\columnwidth]{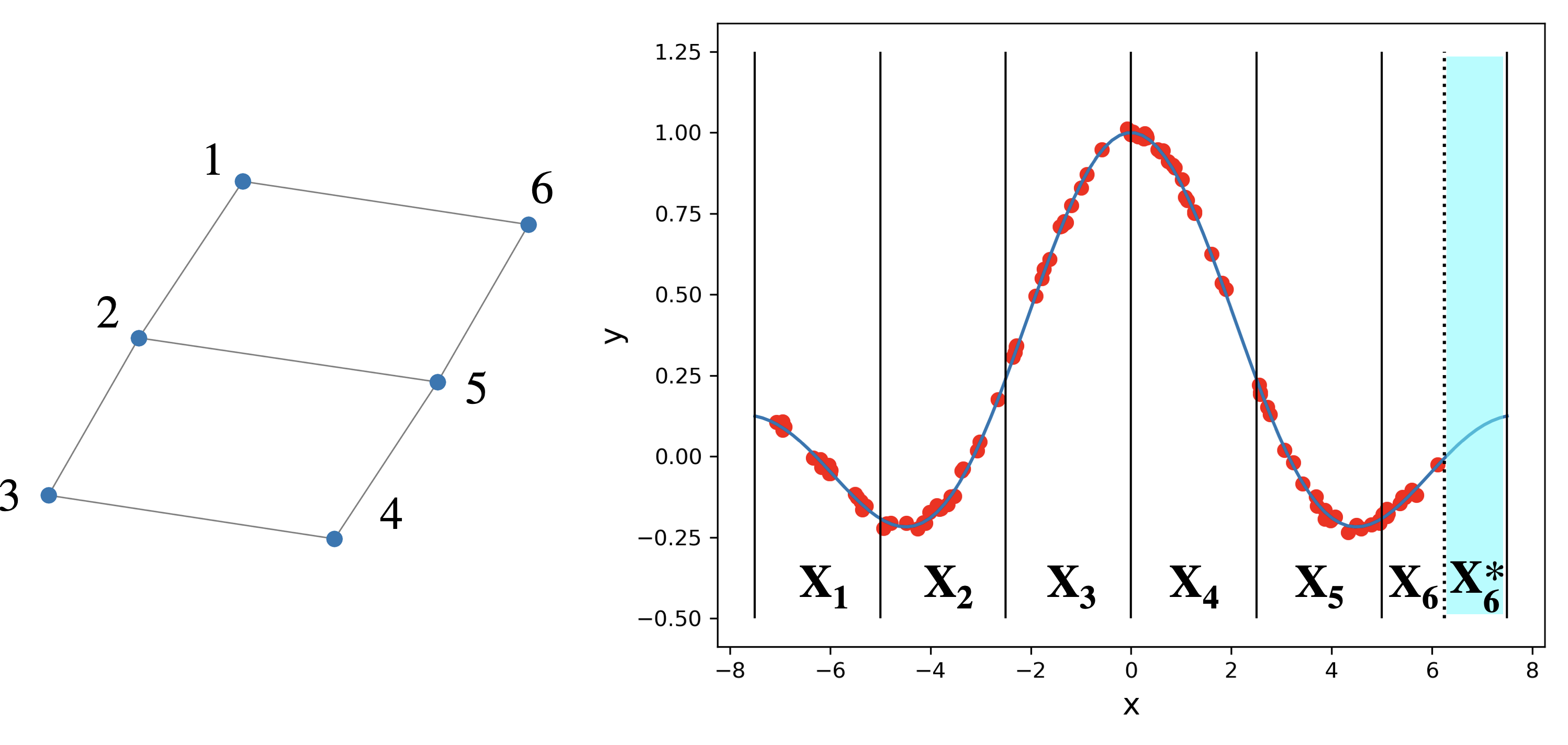}}
  \caption{Graph and data used in the experiments on the synthetic data in Sect.~\ref{sec:synthetic2}. The red dots represent the observed data. The shaded area represents the region to be predicted.}
  \label{fig:synthetic2}
\end{figure}
\subsection{Prediction on Induced Subgraph with Synthetic Data}
\label{sec:synthetic2}
We evaluated performance with synthetic data in the scenario in Sect.~\ref{sec:semi} 
where a graph structure is given and the prediction is performed on the induced subgraph.

The graph structure and the data are shown in Fig.~\ref{fig:synthetic2}.
Each vertex $m$ has the unique space $\bm{X}_m\in\mathbb{R}$ delimited by the vertical line on the right in the figure.
The data $y_i$ are generated from the following function:
\begin{equation}
  y_i = \frac{\sin(x_i)}{x_i} + \epsilon_i,
\end{equation}
where $\epsilon_i\sim\mathcal{N}(0,0.01)$ is the additive Gaussian noise.
The connected vertices in the graph have adjacent input regions and/or symmetric outputs.
The vertices $1$ to $5$ have $N_1=N_2=\ldots=N_5=20$ data points, 
and the vertex $6$ has $N_6=10$ data points, the inputs of which are uniformly and randomly sampled.
We performed prediction with $100$ trials on different randomly chosen $\tilde{T}=10$ test points in $\bm{X}_6^\ast$.

We employed separable kernels with graph kernels in Table~\ref{tab:graphkernels}, 
SoS kernels, and graph PC.
SOGP and ICM were comparison methods that do not include graph information.
The method using polynomial-$d$ \cite{Zhi} was not included in the comparison 
because it cannot be a formulation that utilizes graph information 
for making predictions about a single vertex, as in this problem.
The SE kernel was used as $k(\cdot,\cdot)$ in all methods.
For SoS kernels, we set $Q=2$ and $k_1(\cdot,\cdot)=k_2(\cdot,\cdot)=k(\cdot,\cdot)$, 
and used regularized Laplacian and diffusion as $k_{\mathrm{G},q}(\cdot,\cdot)$.
For the graph PC, we employed the choices of \eqref{eq:ss} and \eqref{eq:pp}, 
and global filtering and graph Mat\'{e}rn-2 with unnormalized graph Laplacian were used, respectively.
We set $\bm{\Lambda}=\ell\bm{I}$.
These kernels were chosen to accommodate both broad and local effects on the graph spectrum.

The performance of the proposed extensions and the baseline methods is summarized in Table~\ref{tab:results1}.
The mean and standard deviation of each metric of the $100$ trials are reported.
From Table~\ref{tab:results1}, when a kernel that could make better use of the graph information was selected, 
it made better predictions than SOGP or ICM that did not use the graph information.
The proposed graph PC achieves the lowest MSE and highest log-likelihood.

\begin{figure}[tb]
  \centering
  \begin{minipage}[b]{0.49\columnwidth}
      \centering
      \includegraphics[width=0.9\columnwidth]{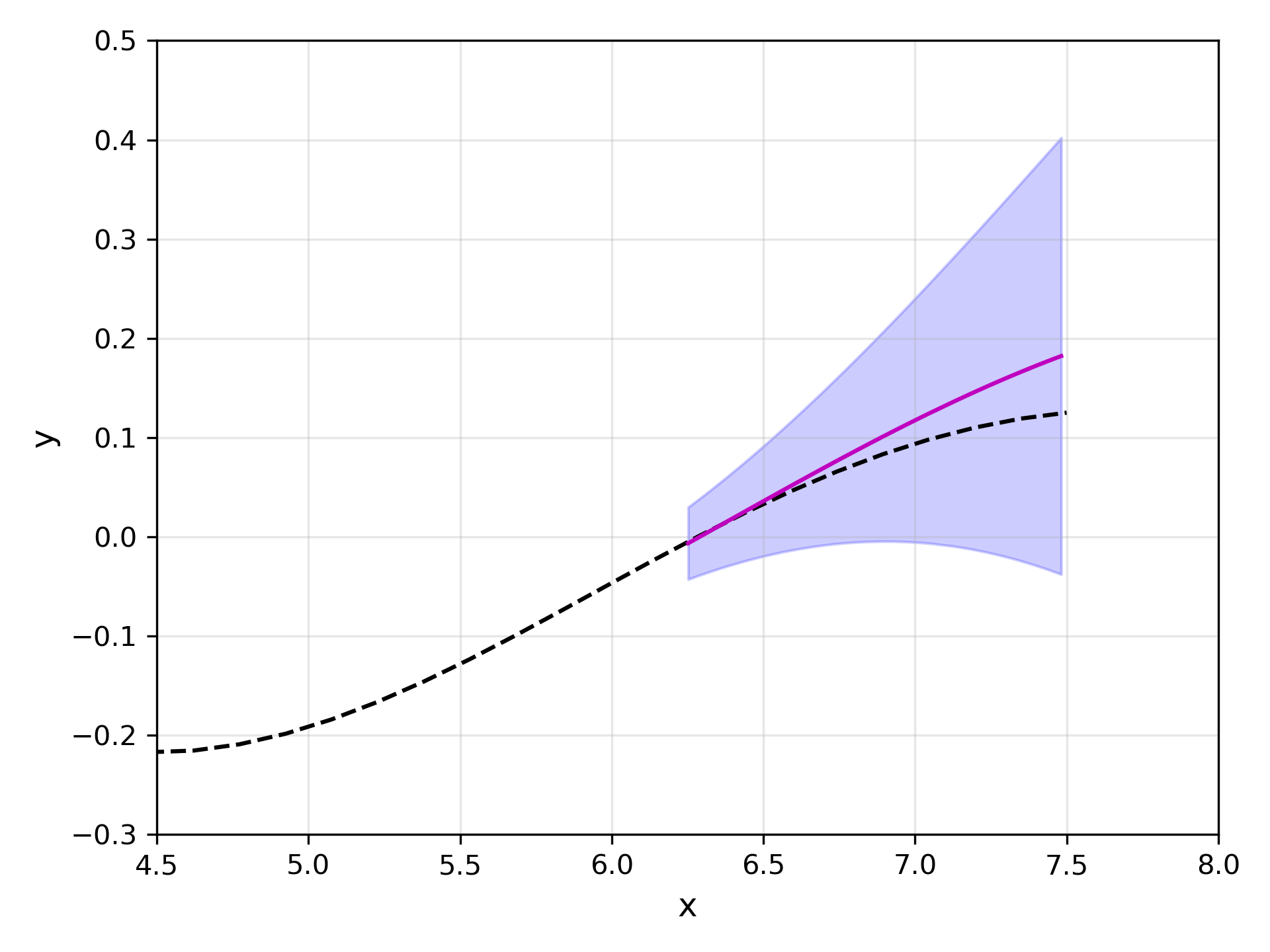}
      \subcaption{SOGP}
      \label{fig:a}
  \end{minipage}
  \begin{minipage}[b]{0.49\columnwidth}
      \centering
      \includegraphics[width=0.9\columnwidth]{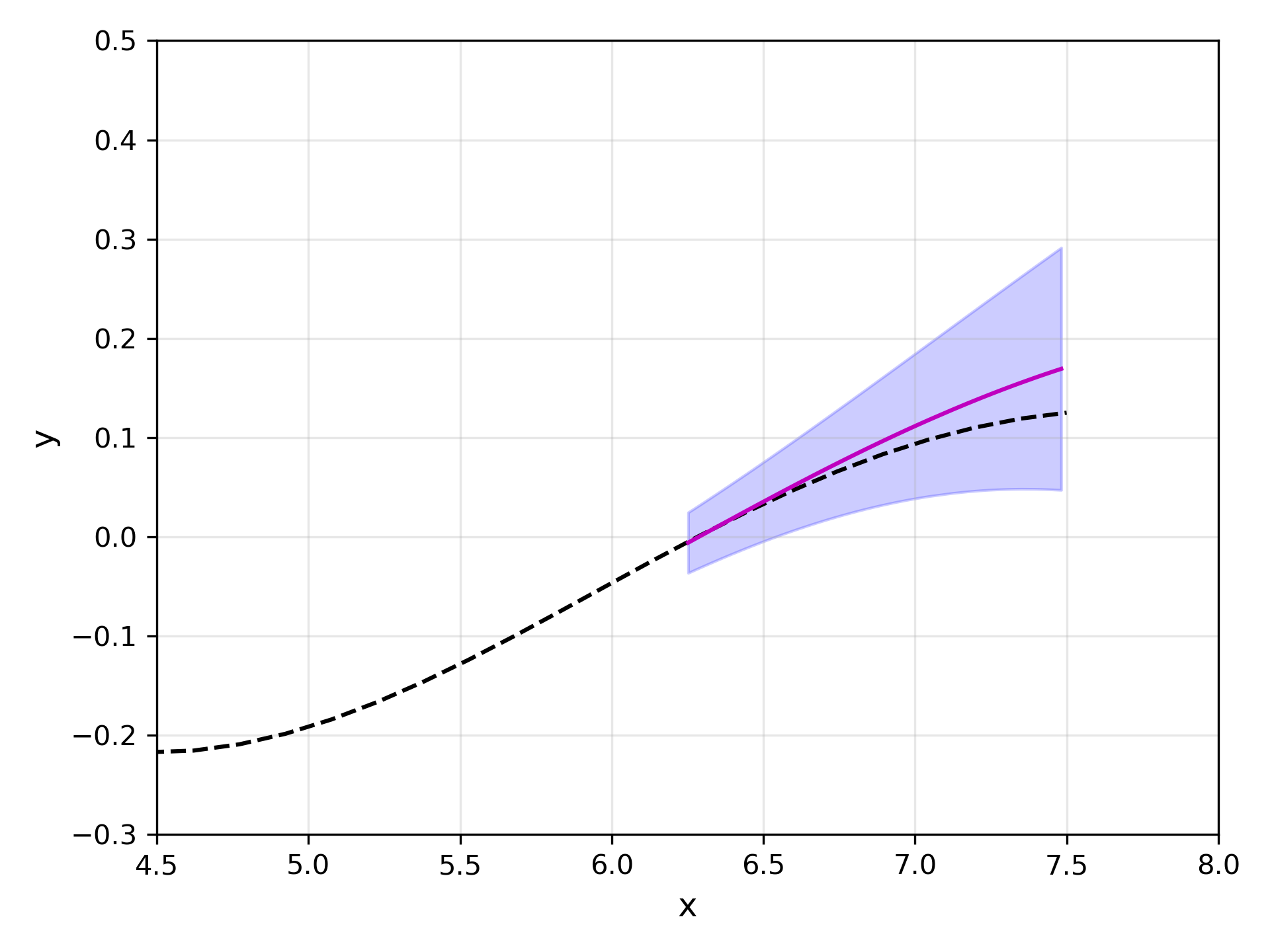}
      \subcaption{ICM}
      \label{fig:b}
  \end{minipage}\\
  \begin{minipage}[b]{0.49\columnwidth}
      \centering
      \includegraphics[width=0.9\columnwidth]{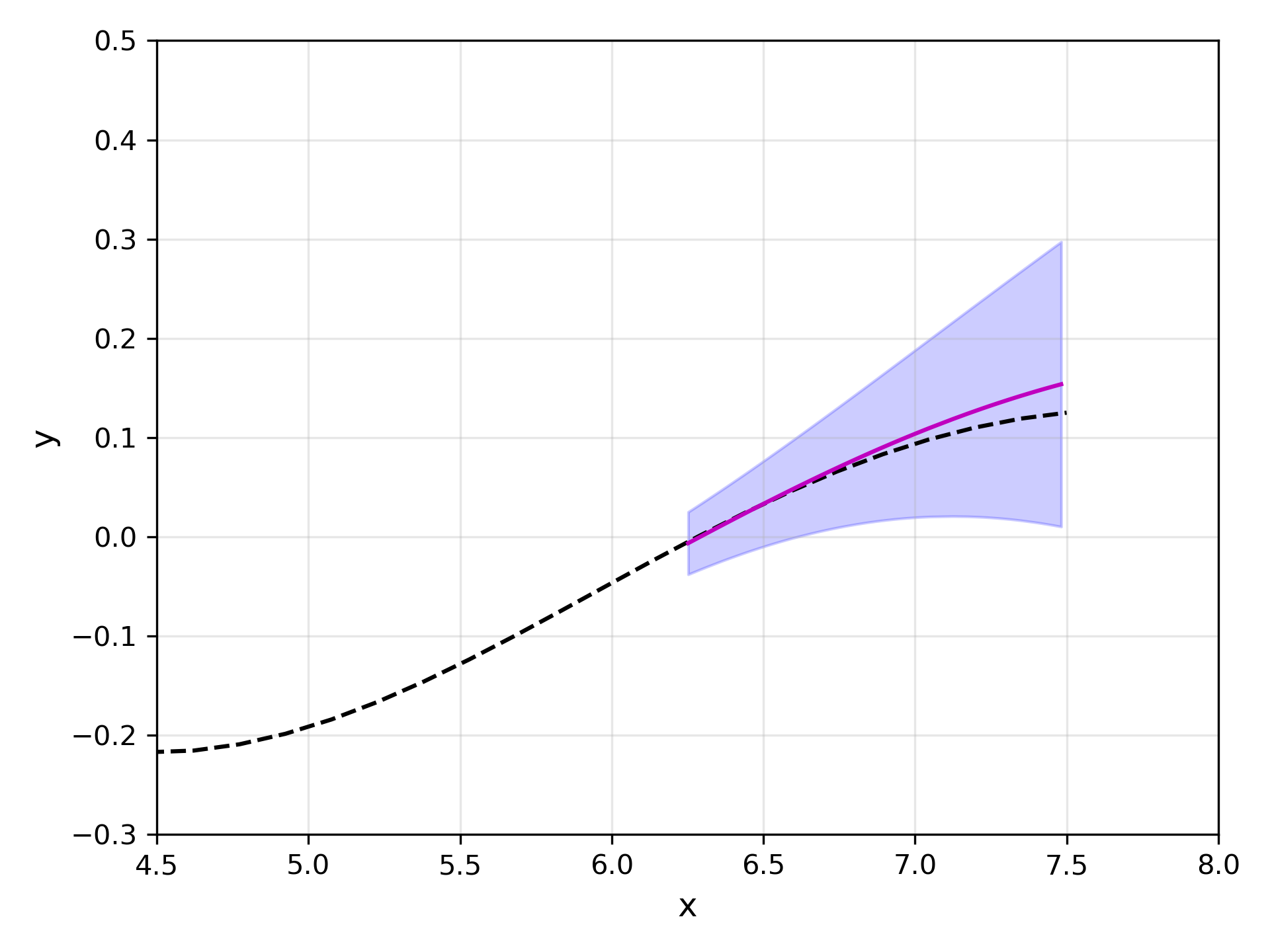}
      \subcaption{Global filtering}
      \label{fig:c}
  \end{minipage}
  \begin{minipage}[b]{0.49\columnwidth}
      \centering
      \includegraphics[width=0.9\columnwidth]{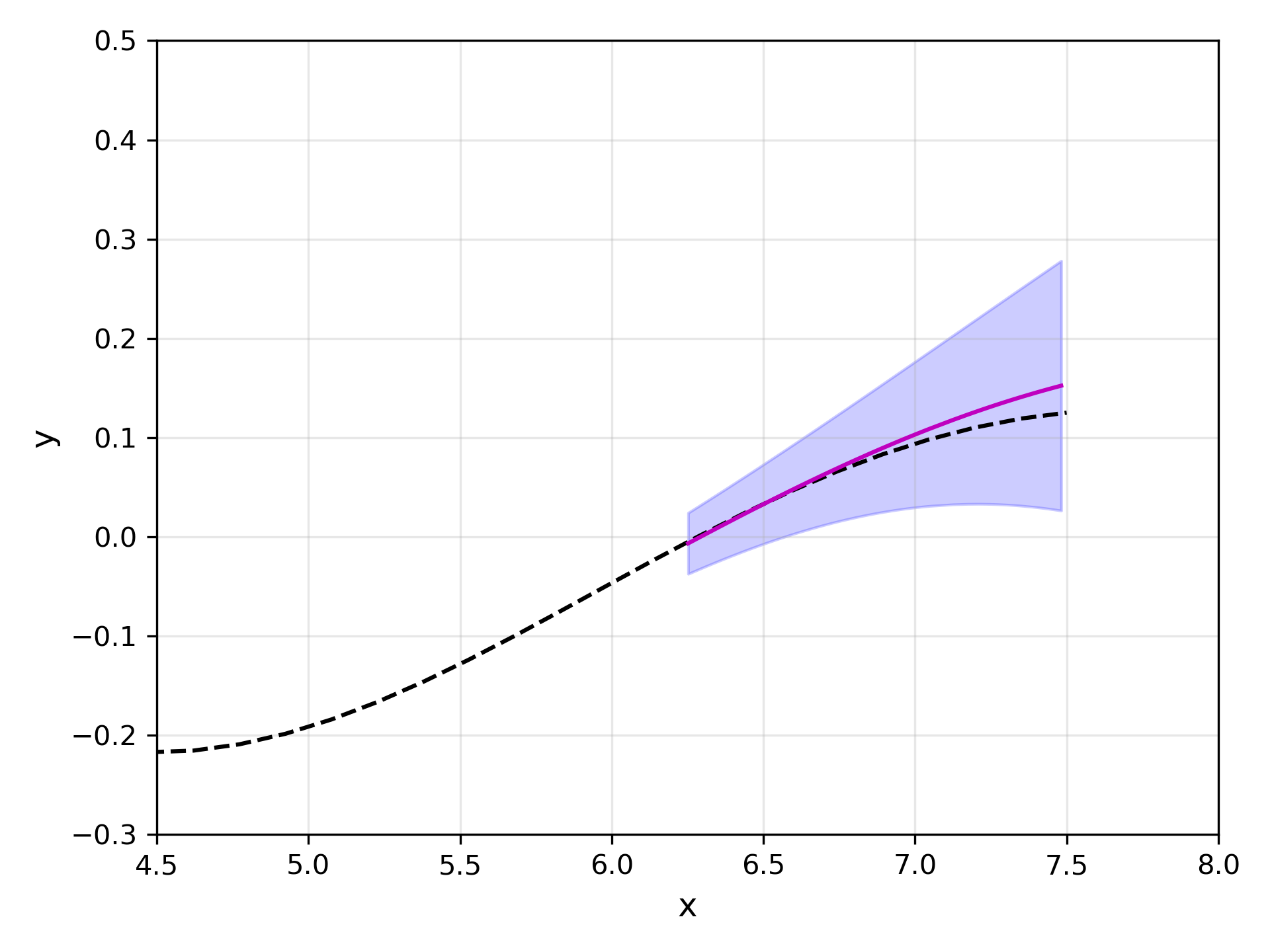}
      \subcaption{Graph PC}
      \label{fig:d}
  \end{minipage}
  \caption{Predicted means (red lines) and 95\% confidence intervals (shaded area) on the synthetic data in Sect.~\ref{sec:synthetic2} for four representative methods.}
  \label{fig:synthetic2_plot}
\end{figure}

Predicted means and 95\% confidence intervals for four representative methods are shown in Fig.~\ref{fig:synthetic2_plot}.
The introduction of graphical information as in Figs.~\ref{fig:c} and \ref{fig:d} led to the predicted mean being closer to the true function 
and the prediction being more reliable.
\begin{table}[t]
  \centering
  \caption{Performance of the proposed formulation on the synthetic data in Sect.~\ref{sec:synthetic2}. PE means the proposed extensions.}
  \label{tab:results1}
  \begin{tabular}{|c|c|c|c|}
    \hline 
    \multicolumn{2}{|c|}{Method} & MSE$(\times 10^{-4}) ~ \blacktriangledown$ & Log-likelihood $\blacktriangle$ \\
    \hline
    \multicolumn{2}{|c|}{SOGP} & $8.327\pm 0.318$ & $2.854\pm 0.0142$ \\
    \multicolumn{2}{|c|}{ICM} & $5.327\pm 0.223$ & $2.945\pm 0.0163$ \\
    \hline
     & Laplacian & $13.62\pm 0.535$ & $2.789\pm 0.0160$ \\
     & Global filtering & $2.620\pm 0.108$ & $2.938\pm 0.0148$ \\
     & Local averaging & $3.344\pm 0.153$ & $2.876\pm 0.0151$ \\
     & Regularized Laplacian & $4.067\pm 0.197$ & $2.896\pm 0.0152$ \\
     & Diffusion & $9.320\pm 0.395$ & $2.891\pm 0.0197$ \\
     & 1-step random walk & $10.65\pm 0.381$ & $2.801\pm 0.0165$ \\
     PE & 3-step random walk & $5.072\pm 0.213$ & $2.867\pm 0.0177$ \\
     & Cosine & $6.106\pm 0.243$ & $2.842\pm 0.0166$ \\
     & Graph Mat\'{e}rn-2 & $4.881\pm 0.210$ & $2.865\pm 0.0207$ \\
     & Graph Mat\'{e}rn-3 & $5.588\pm 0.223$ & $2.854\pm 0.0174$ \\
     & SoS kernels & $3.156\pm 0.143$ & $2.926\pm 0.0174$ \\
     & Graph PC & $\mathbf{2.468\pm 0.115}$ & $\mathbf{2.946\pm 0.0177}$ \\
    \hline
  \end{tabular}
\end{table}

\subsection{Regression with Real Data}
The performance of the proposed extensions were evaluated also with real data 
of isotopic data configuration and symmetric scenario.
The regression problems on the output of all vertices in Sect.~\ref{sec:formulation} were addressed on two datasets, 
functional magnetic resonance imaging (fMRI) \cite{Venkitaraman,fmri} and weather datasets\cite{Venkitaraman,Zhi,weather}.
The construction of the datasets and graph structure were the same as in \cite{Zhi}.

The fMRI dataset consists of the blood-oxygen-level-dependent (BOLD) signals and 
$50$ of the voxels of the cerebellum region in the brain were extracted.
The data of the first $D=10$ voxels were set to $\bar{\bm{X}}\in\mathbb{R}^{10}$ 
and the output $\bm{y}$ was the data of the remaining $M=40$ voxels corresponding to the vertices of the graph.
The graph becomes unconnected due to the extraction but there are no problems in applying the method and the subgraph information can be utilized.
The number of the training and test points was $N_m=21$ and $T_m=25$ for all $m$, respectively.
For the SoS kernels, we set $Q=2$ and evaluated two cases.
One was with the SE kernel as $k_q(\cdot,\cdot)$, and the global filtering and graph Mat\'{e}rn-2 kernels as $k_{\mathrm{G},1}(\cdot,\cdot)$ and $k_{\mathrm{G},2}(\cdot,\cdot)$, respectively. 
The other was with the SE kernels as $k_1(\cdot,\cdot)$, the Mat\'{e}rn-$1/2$ kernel known as Ornstein-Uhlenbeck kernel as $k_2(\cdot,\cdot)$, 
and the graph Mat\'{e}rn-3 kernel as $k_{\mathrm{G},1}(\cdot,\cdot)$, and the polynomial-3 kernel as $k_{\mathrm{G},2}(\cdot,\cdot)$.
For the graph PC, we employed the choices of \eqref{eq:ss} and \eqref{eq:pp}, 
and diffusion and graph Mat\'{e}rn-2 kernels were used, respectively.
We set $\bm{\Lambda}=\ell\bm{I}$.

The weather dataset consists of the temperature measurement in $M=45$ cities in Sweden \cite{weather}.
The problem is to predict the next-day temperature in the cities.
The input data $\bar{\bm{X}}\in\mathbb{R}^{45}$ were the temperature of a day   
and the output $\bm{y}\in\mathbb{R}^{45}$ was the next-day temperature.
The graph was constructed using $k$-nearest neighbour for $k = 10$.
The number of the training and test points was $N_m=10$ and $T_m=6$ for all $m$, respectively.
For the SoS kernels, we set $Q=2$ and evaluated two cases.
One was with the SE kernel as $k_q(\cdot,\cdot)$, and the diffusion and polynomial-3 kernels as $k_{\mathrm{G},1}(\cdot,\cdot)$ and $k_{\mathrm{G},2}(\cdot,\cdot)$, respectively. 
The other was with the SE kernels as $k_1(\cdot,\cdot)$, the Ornstein-Uhlenbeck kernel as $k_2(\cdot,\cdot)$, 
and the diffusion kernel as $k_{\mathrm{G},1}(\cdot,\cdot)$, and the polynomial-3 kernel as $k_{\mathrm{G},2}(\cdot,\cdot)$.
For the graph PC, we employed the choices of \eqref{eq:ss} and \eqref{eq:pp}, 
and diffusion and graph Mat\'{e}rn-2 kernels were used, respectively.
We set $\bm{\Lambda}=\ell\bm{I}$.

The performance of the proposed extensions and the baseline methods is summarized in Table~\ref{tab:results2}.
The mean and standard deviation of each metric of the $10$ trials are reported.
For the weather dataset, 
the mean of the MSE was the best in SOGP but there is not much difference between the methods considering the standard deviations.
In the log-likelihood, however, there were larger differences.
The SoS kernels (SE+OU), one of the proposed extensions, showed the best performance in both datasets.
It should be noted that the number of hyperparameters increased by the SoS is only a several.
\begin{table*}[tb]
  \centering
  \caption{Performance of the proposed extensions and the baseline methods on the real data. }
  \label{tab:results2}
  \begin{tabular}{|c|c||c|c||c|c|}
    \hline 
    \multicolumn{2}{|c||}{} & \multicolumn{2}{c||}{fMRI} & \multicolumn{2}{c|}{Weather} \\
    \cline{3-6}
    \multicolumn{2}{|c||}{Method} & MSE$(\times 10^{-2}) ~ \blacktriangledown$ & Log-likelihood $\blacktriangle$ & MSE$(\times 10^{-1}) ~ \blacktriangledown$ & Log-likelihood $\blacktriangle$ \\
    \hline
    \multicolumn{2}{|c||}{SOGP} & $1.241\pm 0.163$ & $36.25\pm 0.351$ & $\mathbf{2.391\pm 0.334}$ & $-34.28\pm 4.98$ \\
    \multicolumn{2}{|c||}{ICM} & $24.29\pm 3.48$ & $31.27\pm 1.23$ & $2.450\pm 0.372$ & $-17.10\pm 3.26$ \\
    \multicolumn{2}{|c||}{Laplacian} & $13.47\pm 0.285$ & $-11.70\pm 0.268$ & $8.939\pm 0.802$ & $-63.12\pm 2.71$ \\
     \multicolumn{2}{|c||}{Global filtering} & $1.082\pm 0.137$ & $36.37\pm 0.352$ & $2.541\pm 0.405$ & $-11.48\pm 1.93$ \\
     \multicolumn{2}{|c||}{Local averaging} & $1.115\pm 0.143$ & $36.36\pm 0.352$ & $2.538\pm 0.373$ & $-16.93\pm 3.00$ \\
     \multicolumn{2}{|c||}{Regularized Laplacian} & $1.183\pm 0.154$ & $36.29\pm 0.350$ & $2.741\pm 0.472$ & $-12.09\pm 2.45$ \\
     \multicolumn{2}{|c||}{Diffusion} & $1.143\pm 0.149$ & $36.34\pm 0.351$ & $2.536\pm 0.398$ & $-10.69\pm 2.36$ \\
     \multicolumn{2}{|c||}{1-step random walk} & $1.246\pm 0.163$ & $36.25\pm 0.353$ & $2.386\pm 0.331$ & $-34.78\pm 4.99$ \\
     \multicolumn{2}{|c||}{3-step random walk} & $1.128\pm 0.147$ & $36.35\pm 0.353$ & $2.681\pm 0.448$ & $-16.51\pm 3.91$ \\
     \multicolumn{2}{|c||}{Cosine} & $1.136\pm 0.150$ & $36.34\pm 0.356$ & $2.457\pm 0.360$ & $-35.98\pm 6.30$ \\
     \multicolumn{2}{|c||}{Graph Mat\'{e}rn-2} & $1.117\pm 0.145$ & $36.35\pm 0.352$ & $2.634\pm 0.442$ & $-9.902\pm 2.16$ \\
     \multicolumn{2}{|c||}{Graph Mat\'{e}rn-3} & $1.017\pm 0.129$ & $36.45\pm 0.356$ & $2.555\pm 0.414$ & $-9.530\pm 2.07$ \\
     \multicolumn{2}{|c||}{Polynomial-2} & $1.282\pm 0.162$ & $36.09\pm 0.35$ & $2.596\pm 0.436$ & $-9.098\pm 1.91$ \\
     \multicolumn{2}{|c||}{Polynomial-3} & $1.122\pm 0.144$ & $36.34\pm 0.351$ & $2.835\pm 0.502$ & $-8.393\pm 2.22$ \\
     \hline
      & SoS kernels (SE) & $1.082\pm 0.137$ & $36.37\pm 0.352$ & $3.010\pm 0.538$ & $-7.568\pm 2.35$ \\
     PE & SoS kernels (SE+OU) & $\mathbf{0.8154\pm 0.0947}$ & $\mathbf{37.03\pm 0.261}$ & $2.796\pm 0.417$ & $\mathbf{-4.905\pm 1.72}$ \\
     & Graph PC & $1.315\pm 0.171$ & $36.17\pm 0.343$ & $2.937\pm 0.515$ & $-10.28\pm 2.49$ \\
    \hline
  \end{tabular}
\end{table*}

\section{Conclusions}
\label{sec:conc}
This paper formulated regression models for graph-structured data on the basis of multi-output Gaussian processes.
The proposed formulation allows for the effective use of graph information for a wide range of applications 
and enables the removal of restrictions on data configurations, model selection, and inference scenarios that existed in conventional methods.
The proposed formulation is available for heterotopic data and asymmetric scenario, 
allows the use of flexible kernel selection including the SoS kernels and process convolution, 
and is applicable to a extended inference scenario such as missing values estimation.
This paper also provided a novel kernel design for graph-structured data named graph PC.
Results of computer experiments showed that the extensions enabled by the proposed formulation perform well on both synthetic and real graph structure data.

Models based on multi-output Gaussian processes are flexible but suffer from high computational costs, 
even though it is serious also in SOGP.
The inverse calculation in terms of the covariance matrix $\bm{K}_{\mathcal{M}}(\bm{X})$ requires $O(N^3)$ time complexity, which is computationally expensive for large-scale graph-structured data.
In future work, we will address this issue 
by developing sparsification of covariance matrices taking into account the graph structure.

\section*{Acknowledgments}
This work was supported by 
JSPS KAKENHI Grant-in-Aid for Young Scientists Grant Number JP23K13334 (to A. Nakai-Kasai).

 




\vspace{11pt}
\begin{IEEEbiography}[{\includegraphics[width=1in,height=1.25in,clip,keepaspectratio]{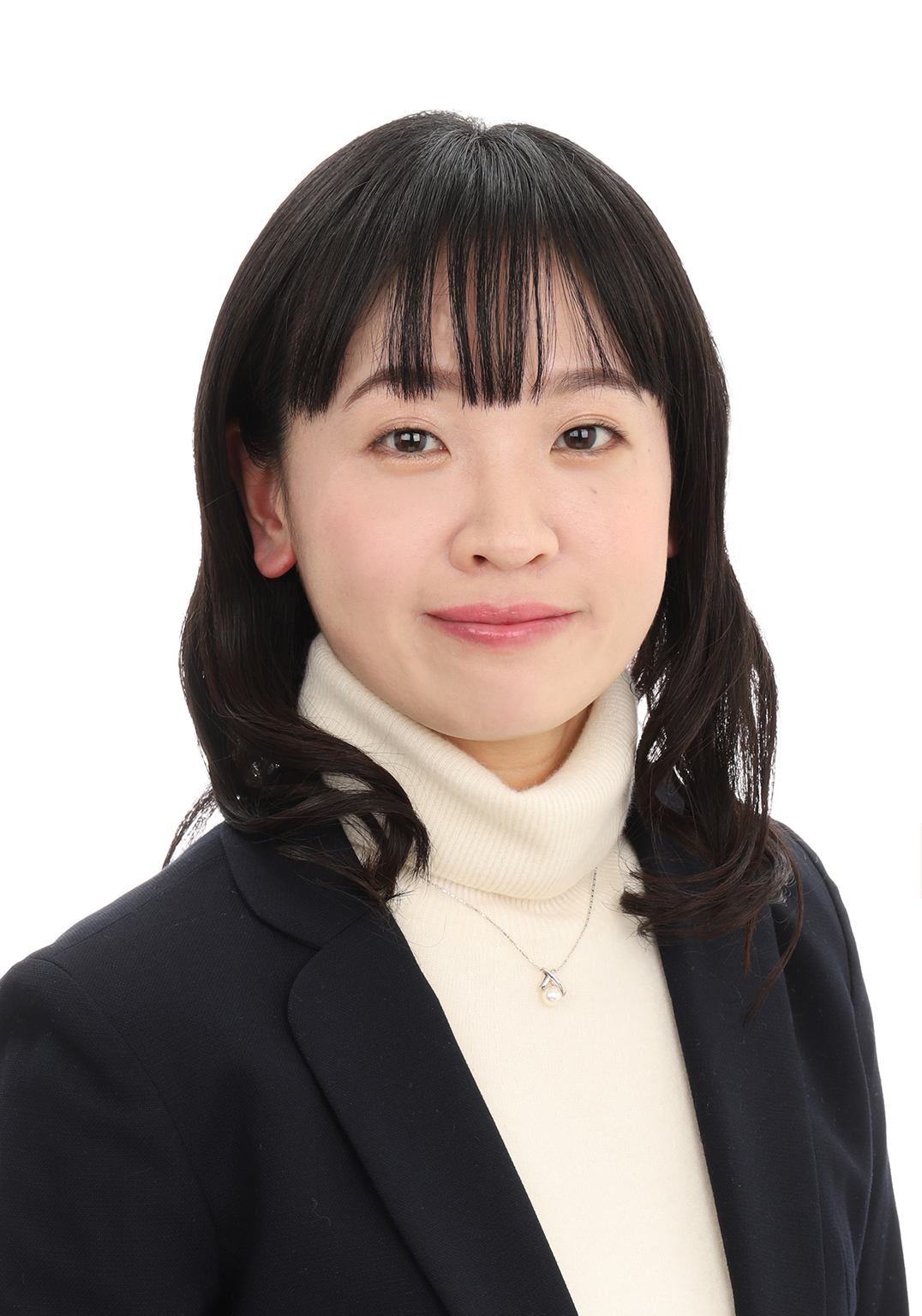}}]{Ayano Nakai-Kasai}
    received the bachelor's degree 
    in engineering, the master's degree in informatics,
    and Ph.D. degree in informatics from Kyoto University, Kyoto, Japan, in 2016, 2018, and 2021,
    respectively. She is currently an Assistant Professor
    at Graduate School of Engineering, Nagoya Institute of Technology. Her research interests include
    signal processing, wireless communication, and machine learning. She received the Young Researchers'
    Award from the Institute of Electronics, Information 
    and Communication Engineers in 2018 and APSIPA 
    ASC 2019 Best Special Session Paper Nomination Award. She is a member
    of IEEE and IEICE.
\end{IEEEbiography}
\begin{IEEEbiography}[{\includegraphics[width=1in,height=1.25in,clip,keepaspectratio]{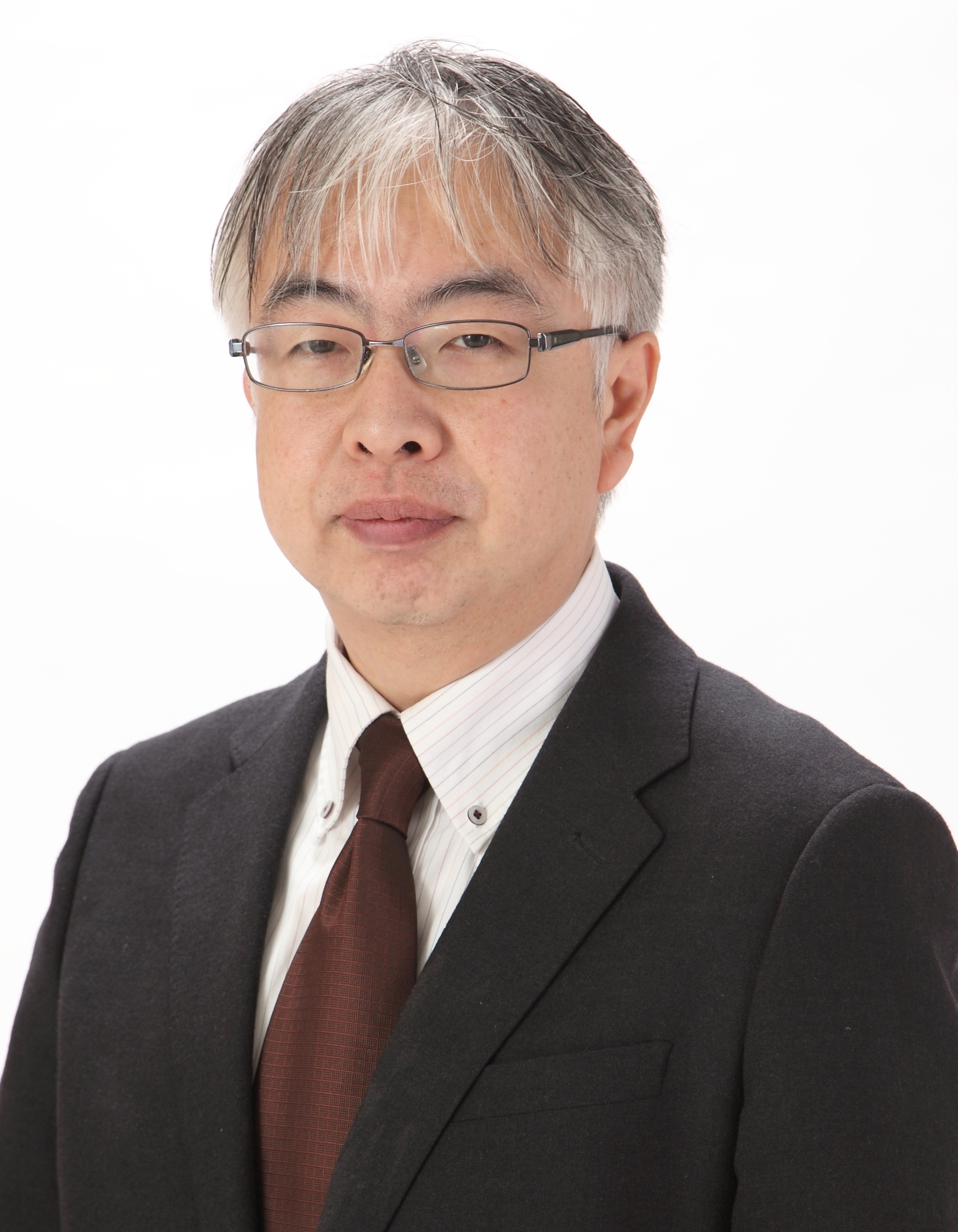}}]{Tadashi Wadayama}
    (M'96) was born in Kyoto, Japan, on May 9,1968.
    He received the B.E., the M.E., and the D.E. degrees from Kyoto Institute of Technology in 1991, 1993 and 1997, respectively.
    On 1995, he started to work with Faculty of Computer Science and System Engineering, Okayama Prefectural University as a research associate.
    From April 1999 to March 2000, he stayed in Institute of Experimental Mathematics, Essen University (Germany) as a visiting researcher.
    On 2004, he moved to Nagoya Institute of Technology as an associate professor. Since 2010, he has been a full professor of Nagoya Institute of Technology.
    His research interests are in coding theory, information theory, and signal processing for wireless communications.
    He is a member of IEEE and a senior member of IEICE.
\end{IEEEbiography}

\vfill

\end{document}